\documentclass[10pt,journal,compsoc]{IEEEtran}

\usepackage{graphicx}
\usepackage[cmex10]{amsmath}
\usepackage{color}
\usepackage{booktabs}
\usepackage{epsfig}
\usepackage{graphicx}
\usepackage{amsmath}
\usepackage{amssymb}
\usepackage{array}
\usepackage{url}
\usepackage{multirow}
\usepackage{paralist}
\usepackage{bbding}
\usepackage[pagebackref=true,breaklinks=true,colorlinks,citecolor=blue,linkcolor=blue,bookmarks=false]{hyperref}
\usepackage{algorithm}
\usepackage{algorithmic}

\usepackage[nocompress]{cite}

\renewcommand{\raggedright}{\leftskip=0pt \rightskip=0pt plus 0cm}

\begin{document}

\title{Self-Supervised Deep Blind Video Super-Resolution}

\author{%
	Haoran Bai
	and~Jinshan Pan% <-this % stops a space
	\IEEEcompsocitemizethanks{
		\IEEEcompsocthanksitem The authors are with the School of Computer Science and Engineering, Nanjing University of Science and Technology, Nanjing, 210094, China.
		%\protect\\
		% note need leading \protect in front of \\ to get a newline within \thanks as
		% \\ is fragile and will error, could use \hfil\break instead.
		E-mail: \{baihaoran@njust.edu.cn, sdluran@gmail.com.\}
		%\IEEEcompsocthanksitem Corresponding author: Jinshan Pan
	}% <-this % stops an unwanted space
}

\IEEEtitleabstractindextext{%
\begin{abstract}
\raggedright{
	Existing deep learning-based video super-resolution (SR) methods usually depend on the supervised learning approach, where the training data is usually generated by the blurring operation with known or predefined kernels (e.g., Bicubic kernel) followed by a decimation operation. However, this does not hold for real applications as the degradation process is complex and cannot be approximated by these idea cases well. Moreover, obtaining high-resolution (HR) videos and the corresponding low-resolution (LR) ones in real-world scenarios is difficult. To overcome these problems, we propose a self-supervised learning method to solve the blind video SR problem, which simultaneously estimates blur kernels and HR videos from the LR videos. As directly using LR videos as supervision usually leads to trivial solutions, we develop a simple and effective method to generate auxiliary paired data from original LR videos according to the image formation of video SR, so that the networks can be better constrained by the generated paired data for both blur kernel estimation and latent HR video restoration. In addition, we introduce an optical flow estimation module to exploit the information from adjacent frames for HR video restoration. Experiments show that our method performs favorably against state-of-the-art ones on benchmarks and real-world videos.
}
\end{abstract}

% Note that keywords are not normally used for peerreview papers.
}

% make the title area
\maketitle

%\IEEEdisplaynontitleabstractindextext
%\IEEEpeerreviewmaketitle

%-------------------------------------------------------
\IEEEraisesectionheading{\section{Introduction}\label{sec:introduction}}

%--------------------------------
\begin{figure*}[!t]\footnotesize
	\begin{center}
		\begin{tabular}{ccccccc}
			\multicolumn{3}{c}{\multirow{5}*[60.5pt]{\includegraphics[width=0.410\linewidth, height = 0.281\linewidth]{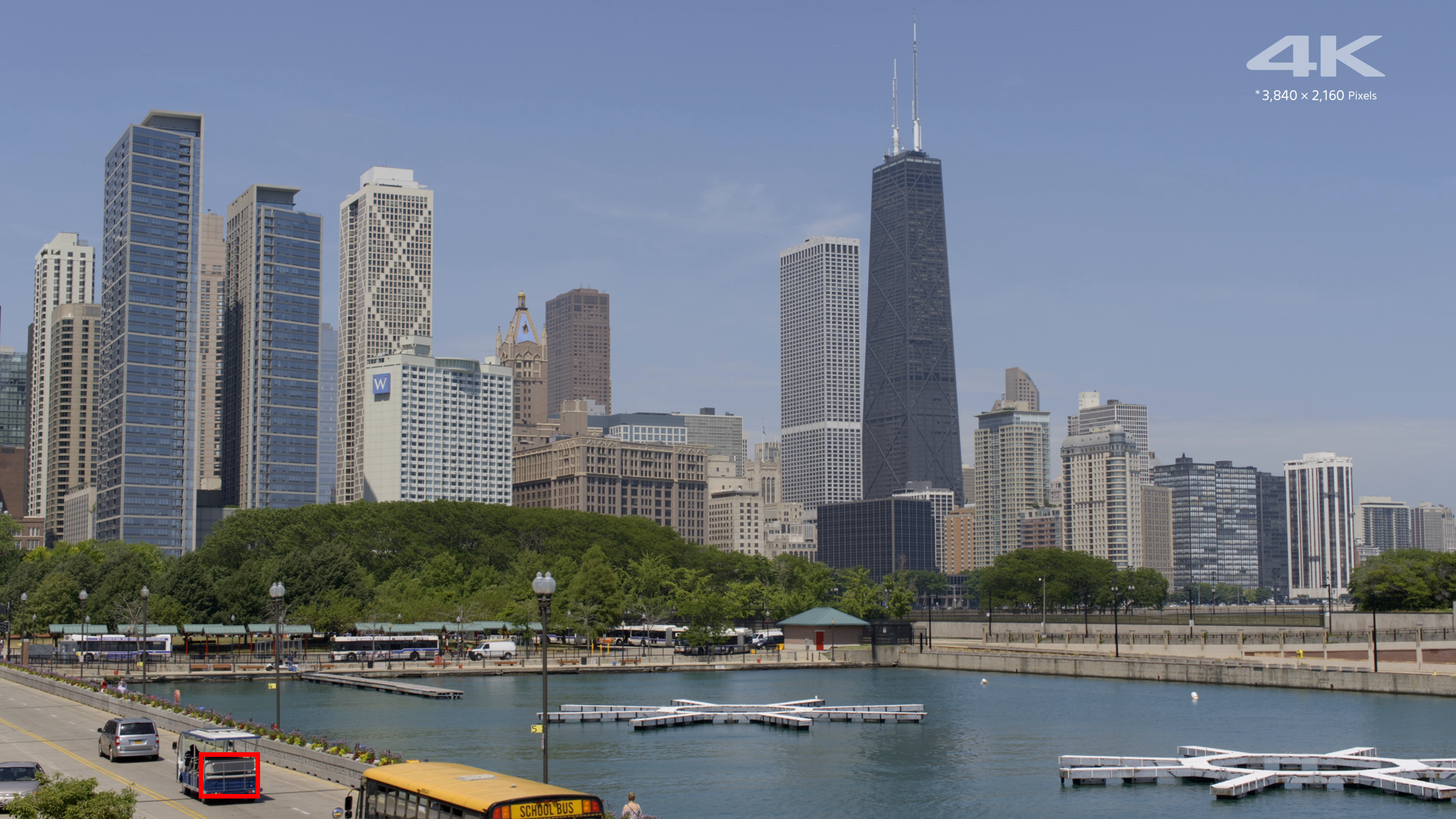}}}&\hspace{-4.5mm}
			\includegraphics[width=0.14\linewidth, height = 0.129\linewidth]{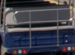} &\hspace{-4.5mm}
			\includegraphics[width=0.14\linewidth, height = 0.129\linewidth]{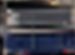} &\hspace{-4.5mm}
			\includegraphics[width=0.14\linewidth, height = 0.129\linewidth]{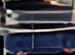} &\hspace{-4.5mm}
			\includegraphics[width=0.14\linewidth, height = 0.129\linewidth]{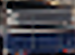} \\
			\multicolumn{3}{c}{~} &\hspace{-4.5mm}  (b) HR patch &\hspace{-4.5mm}  (c) Bicubic &\hspace{-4.5mm}  (d) IKC~\cite{IKC} &\hspace{-4.5mm}  (e) ZSSR~\cite{zssr} \\
			\multicolumn{3}{c}{~} & \hspace{-4.5mm}
			\includegraphics[width=0.14\linewidth, height = 0.129\linewidth]{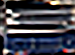} & \hspace{-4.5mm}
			\includegraphics[width=0.14\linewidth, height = 0.129\linewidth]{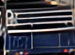} & \hspace{-4.5mm}
			\includegraphics[width=0.14\linewidth, height = 0.129\linewidth]{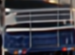} & \hspace{-4.5mm}
			\includegraphics[width=0.14\linewidth, height = 0.129\linewidth]{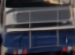} \\
			\multicolumn{3}{c}{\hspace{-4.5mm} (a) Ground truth HR frame} &  \hspace{-4.5mm} (f) KernelGAN~\cite{kernelgan} &\hspace{-4.5mm}  (g) RBPN~\cite{VDBPN/cvpr19} &\hspace{-4.5mm}  (h) EDVR~\cite{edvr} & \hspace{-4.5mm} (i) Ours\\
		\end{tabular}
	\end{center}
	\caption{Video super-resolution results ($\times 4$). We develop an effective self-supervised video SR method that does not require the corresponding ground truth HR videos or any other additional HR videos as the supervision, and it can recover more accurate structural details than those by the supervised learning-based methods.}
	\label{fig: teaser}
\end{figure*}
%--------------------------------

\IEEEPARstart{W}{ith} the rapid development of high definition devices, visualizing the videos generated by some low-resolution (LR) imaging devices on these high definition devices usually leads to significant aliasing and blur effect.
Thus, it is of great interest to develop an effective algorithm to super-resolve videos for better visualization on the high definition devices.

The goal of video super-resolution (SR) is to infer the latent high-resolution (HR) videos from given LR ones.
The degradation process of the video SR problem is usually modeled as~\cite{Bayesian/vsr/pami/LiuS14}:
\begin{equation}
	\mathrm{y}_j =\mathbf{S}\mathbf{K}_j\mathbf{F}_{i\to j}\mathrm{x}_i + n, j = i-N, i-N+1,...,i+N,
	\label{eq: sr-formation-convolution}
\end{equation}
where $\mathrm{y}_j$, $\mathrm{x}_i$, and $n$ denote the $j$-th LR frame, $i$-th HR frame, and noise, respectively; $\mathbf{S}$ and $\mathbf{K}_j$ denote the downsampling and blurring matrix w.r.t. $S$ and $K_j$; $\mathbf{F}_{i\to j}$ denotes the warping matrix which warps $\mathrm{x}_i$ to the $j$-th frame.
Video SR is a highly ill-posed problem as the latent HR frame, blurring matrix, and warping matrix are unknown.

Some recent significant advance has been achieved due to the use of kinds of deep convolution neural networks (CNNs).
However, the ground truth HR videos are always required to train deep models, which cannot be easily satisfied in real applications.
To constrain the deep model training, most video SR methods~\cite{huang/videosr/pami18,renjialiao/cvpr15,Kappeler/tmi16,VESPCN,ding/liu/iccv17,taoxin/iccv17,DUF,edvr,TDAN/CVPR2020} assume that the blur kernels are known or predefined (e.g., Bicubic kernel) for synthesizing the paired datasets.
Although they can achieve state-of-the-art results on existing benchmark datasets, as shown in Figure~\ref{fig: teaser}(g) and (h), the deep models trained on the synthetic datasets generated by predefined blur kernels cannot be generalized well on real videos because blur kernels in real applications are much more complex.

To generate realistic blur kernels for the SR problem, several methods develop effective deep models to estimate blur kernels from input LR images~\cite{kernelgan,IKC}.
Another kind of methods develop effective unpaired learning~\cite{PseudoSR/cvpr20} and zero-shot learning methods~\cite{zssr} to solve the SR problem.
These methods achieve decent performance on real-world applications.
However, most of these methods are designed for the single image SR problem, and few are developed for the video SR problem.
Directly applying them to the video SR problem can not produce convincing results as shown in Figure~\ref{fig: teaser}(d)-(f).

In this paper, we present an effective video SR algorithm based on self-supervised learning, which can simultaneously estimate the blur kernels and the latent HR frames based on deep CNNs.
Instead of synthesizing unpaired datasets~\cite{PseudoSR/cvpr20} which usually requires sophisticated network designs, our algorithm only explores the information from the LR input videos for HR ones restoration.
We first develop deep CNN models to estimate blur kernels and latent HR videos from LR input videos.
Then, we use the original LR input videos to constrain the regenerated LR videos, which are generated by the estimated blur kernels and latent HR videos according to the image formation of video SR, so that the deep models can be learned.
However, directly using LR input videos as the supervision of network training usually leads to trivial solutions.
To address this issue, we first explore the sparse property of blur kernels to constrain the blur kernel estimation network, and then develop a simple and effective method for generating auxiliary paired data from the original LR input videos based on the image formation of video SR, so that the networks can be better constrained by both the generated paired data and sparse property constraint for both blur kernel estimation and latent HR video restoration.
In addition, we introduce an optical flow estimation module to exploit the information from adjacent frames for better HR video restoration.
By training the proposed algorithm in an end-to-end fashion, we show that it performs favorably against state-of-the-art methods on benchmark datasets and real-world videos.
To the best of our knowledge, this is the first algorithm that develops a self-supervised learning method for blind video SR.
Figure~\ref{fig: teaser} shows one super-resolved frame, where the proposed method generates the results with correct structural details than both single and video SR methods.

The main contributions are summarized as follows:
\begin{compactitem}
	\item We propose an effective self-supervised learning algorithm for video SR that does not require any paired or unpaired datasets as the supervision.
	\item To constrain the deep models for video SR, we develop a simple and effective method to generate auxiliary paired data from the original LR input videos according to the image formation of video SR.
	\item We train our method in an end-to-end manner and show that it generates favorable results on both benchmark datasets and real-world videos. To the best of our knowledge, this is the first self-supervised learning-based algorithm for blind video SR.
\end{compactitem}

%-------------------------------------------------------
\section{Related Work}

\subsection{Video SR based on known blur kernels}

Due to the ill-posed nature of the video SR problem, developing kinds of priors or regularization has been the focus of much in the past decade, and significant progress has been made~\cite{IBP,vsr/prior/Richard96,shan/sr/sa08,vsr/prior/Milanfar/tip09}.
However, using priors usually leads to complex optimization problems which are difficult to solve.
Motivated by the success of the deep CNNs in single image SR~\cite{SRCNN/ECCV,VDSR,RCAN}, recent methods usually use deep CNNs to solve the video SR problem with motion compensation~\cite{huang/videosr/pami18,renjialiao/cvpr15,Kappeler/tmi16,VESPCN,ding/liu/iccv17,taoxin/iccv17}.
For example, Caballero et al.~\cite{VESPCN} propose an effective upsampling and motion compensation method based on the single image SR~\cite{SRCNN/ECCV} for real-time video SR.
Tao et al.~\cite{taoxin/iccv17} propose an effective sub-pixel motion compensation layer based on the image formation model of the video SR, which is able to restore structural details.
Instead of using explicit motion compensation, Jo et al.~\cite{DUF} learn dynamic upsampling filters and a residual image for effective to restore HR videos.
To better explore the useful information from adjacent frames, the temporal group attention~\cite{TGA/CVPR2020} and deformable alignment network~\cite{edvr,TDAN/CVPR2020} have been proposed.
Although these aforementioned methods achieve decent performance, they usually assume that the blur kernels are known or predefined (e.g., Bicubic kernels).
However, the blur kernels for real images are much more complicated. Thus, these methods cannot be directly applied to real-world applications.

\subsection{Blind Video SR}

Instead of assuming the blur kernels are known, several methods estimate blur kernels from given LR videos.
In~\cite{Bayesian/vsr/pami/LiuS14}, Liu and Sun develop an effective Bayesian adaptive video SR algorithm, where the blur kernels are directly estimated from given LR videos in a Bayesian framework.
Ma et al.~\cite{ziyangma/CVPR15} estimate blur kernels to super-resolve the blurry LR videos.
Although decent performance has been achieved, these methods usually need to solve complex optimization problems, and the performance is limited to the hand-crafted priors.

Instead of using hand-crafted priors, several methods develop deep CNNs to estimate blur kernels for single image SR~\cite{kernelgan,IKC}.
These methods have been shown better results than the hand-crafted prior-based methods~\cite{Tomer/blindsr/iccv13,Bayesian/vsr/pami/LiuS14,ziyangma/CVPR15}.
However, they are designed for single image SR which cannot be applied to video SR.
In~\cite{pan2021deep}, Pan et al. propose a blind video SR method that simultaneously estimates blur kernels and latent HR frames and develop an image deconvolution method for generating sharp intermediate frames to guide the latent HR frames restoration.
Although this method can achieve decent results, it still depends on the supervised learning approach and requires the ground truth HR videos for supervision, which cannot be easily satisfied in real applications.

\subsection{Self-supervised learning-based methods}

Self-supervised learning has been widely developed to solve the image restoration problem (e.g., image denoising~\cite{noisetonoise/icml18}) when the paired training data is not available.
In the SR problem, Bulat et al.~\cite{GAN/data/eccv18} first use a GAN to synthesize paired training datasets and then use the paired training datasets as the supervision for image SR.
Maeda~\cite{PseudoSR/cvpr20} develops an effective unpaired image SR algorithm based on a pseudo-supervision in a unified framework.
These methods perform well on real-world applications. However, they are developed for single image SR, which cannot be directly extended to the video SR problem.

Different from these methods, we develop an effective self-supervised learning method for blind video SR, where the blur kernels and the HR videos are estimated simultaneously so that the HR videos can be better restored.
%

%-------------------------------------------------------
\section{Proposed Algorithm}

%--------------------------------
\begin{figure*}[!t]\footnotesize
	\begin{center}
		\begin{tabular}{c}
			\includegraphics[width = 0.98\linewidth]{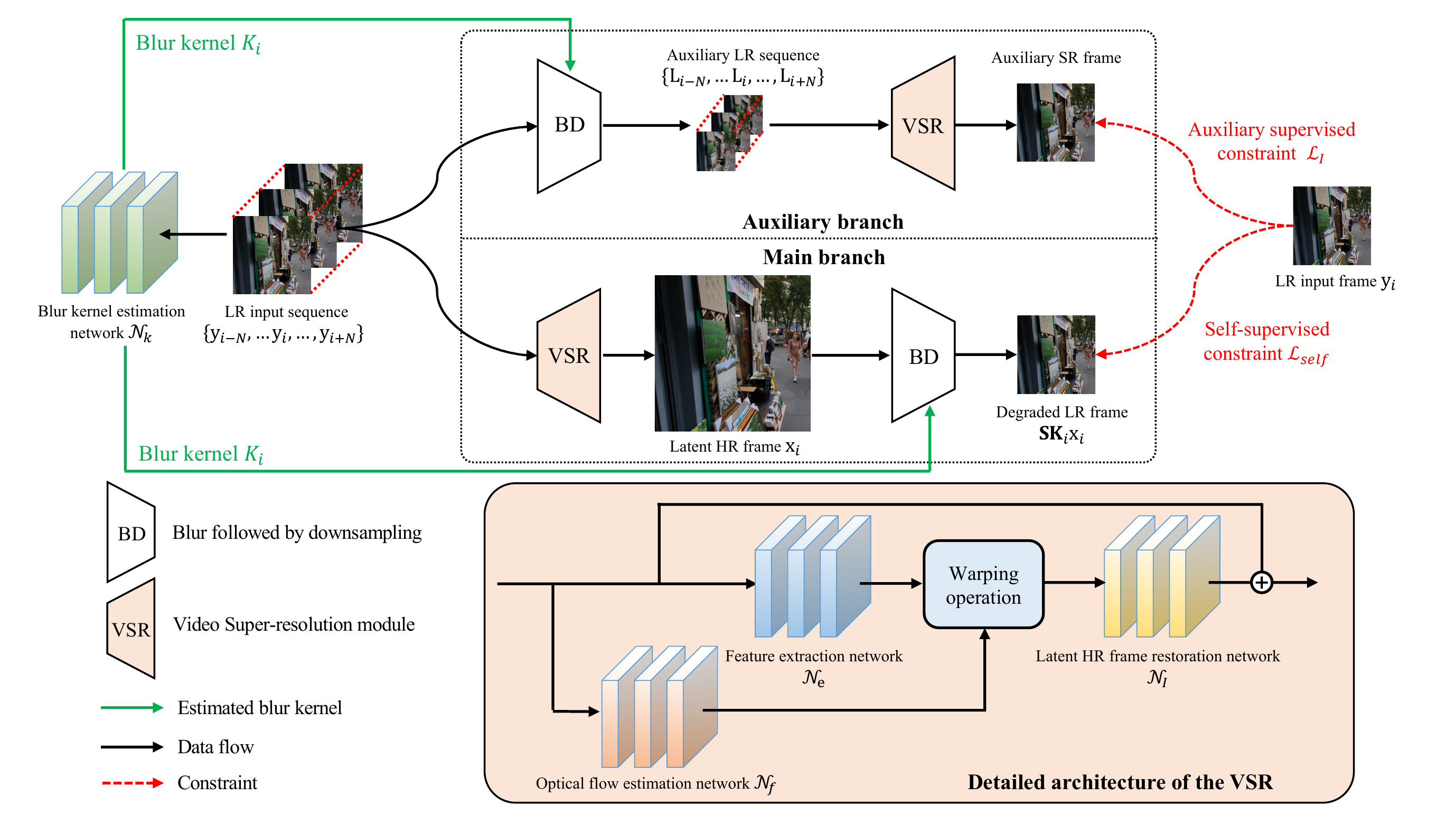}\\
		\end{tabular}
	\end{center}
	\caption{%
		An overview of the proposed method.
		The proposed self-supervised learning-based deep CNN model contains two branches.
		The main branch is used to estimate blur kernel, the optical flow, and latent HR frame under the self-supervision of the LR input frame.
		The auxiliary branch uses the auxiliary paired data, which is generated based on the LR input frames and estimated blur kernel from the main branch, to constrain the network training for the optical flow and latent HR frame.
		The video super-resolution module (VSR) in these two branches share the same network parameters.
		All the branches are jointly trained in an end-to-end manner based on the self-supervised learning method. Please refer to the main content for details.
	}
	\label{fig: flow-chart}
\end{figure*}
%--------------------------------

%--------------------------------
\begin{figure*}[!t]\footnotesize
	\begin{center}
		\begin{tabular}{c}
			\includegraphics[width = 0.96\linewidth]{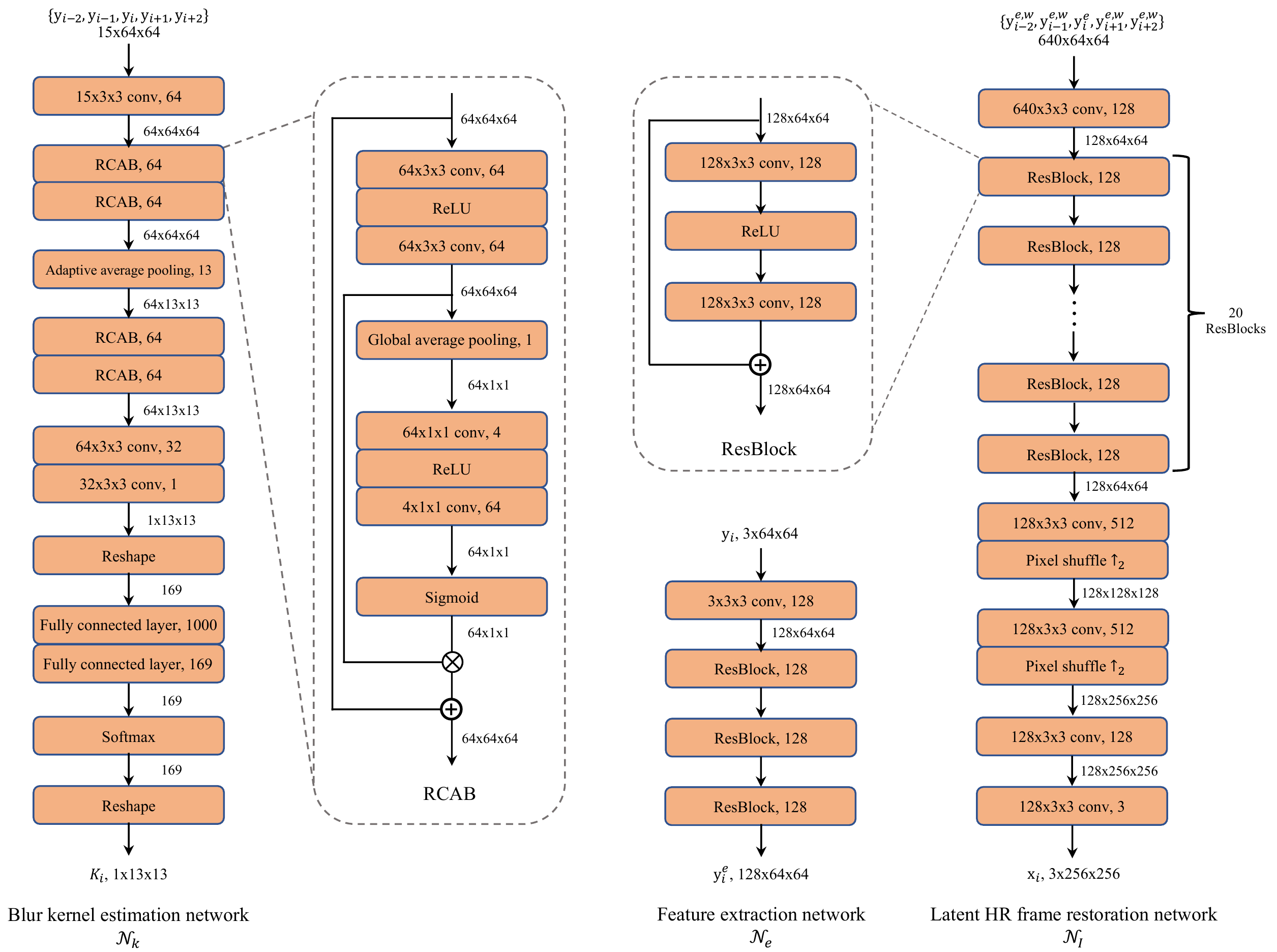}\\
		\end{tabular}
	\end{center}
	\caption{The detailed network configurations of the blur kernel estimation network $\mathcal{N}_k$, the feature extraction network $\mathcal{N}_e$, and the latent HR frame restoration network $\mathcal{N}_I$.}
	\label{fig: network-configurations}
\end{figure*}
%--------------------------------

Given the LR sequence $\{\mathrm{y}_i\}$, the proposed method aims to estimate the HR sequence $\{\mathrm{x}_i\}$ without any supervision of the ground truth HR sequences.
For simplicity, we assume that the latent HR frame $\mathrm{x}_i$ is estimated by $\{\mathrm{y}_{i-N},..., \mathrm{y}_{i-1}, \mathrm{y}_i, \mathrm{y}_{i+1}, ..., \mathrm{y}_{i+N}\}$.
Based on the image formation model~\eqref{eq: sr-formation-convolution}, recovering the HR frame $\mathrm{x}_i$ needs to estimate the blur kernel and warping matrix (w.r.t. optical flow).
Therefore, we develop an effective self-supervised learning approach so that the blur kernels, the optical flow, and the latent HR frames can be simultaneously estimated without any HR sequence supervision.
Based on the proposed self-supervised learning approach, the deep CNN model is designed as two branches.
The main branch is used to estimate blur kernel, the optical flow, and latent HR frame, and the auxiliary branch uses the auxiliary paired training data which is generated based on the LR input frames and estimated blur kernel from the main branch to constrain the network training for the optical flow and latent HR frame.
All the branches are jointly trained in an end-to-end manner based on the self-supervised learning method.
Figure~\ref{fig: flow-chart} shows an overview of the proposed algorithm.
In the following, we first introduce the network designs about the blur kernel estimation, optical flow estimation, and latent HR frame restoration and then present the proposed self-supervised learning approach to solve the blind video SR problem.

\subsection{Blur kernel estimation}
\label{ssec: Blur kernel estimation}

The blur kernel estimation module aims to estimate the blur kernel $K_i$ from $\{\mathrm{y}_{i-N},..., \mathrm{y}_{i-1}, \mathrm{y}_i, \mathrm{y}_{i+1}, ..., \mathrm{y}_{i+N}\}$.
Let $\mathcal{N}_k$ denotes the blur kernel estimation network, and we estimate blur kernel $K_i$ by:
\begin{equation}
	K_i  = \mathcal{N}_k(\mathcal{C}[\mathrm{y}_{i-N},..., \mathrm{y}_{i-1}, \mathrm{y}_i, \mathrm{y}_{i+1}, ..., \mathrm{y}_{i+N}]),
	\label{eq: self-supervised-loss-hr}
\end{equation}
where $\mathcal{C}[\cdot]$ denotes a concatenation operation.
For the network $\mathcal{N}_k$, we first adopt a convolutional network with a pooling operation to extract the information of the blur kernel from the input.
Then, we apply two fully connected layers followed by a softmax activation function to obtain the blur kernel and ensure that the sum of all elements is 1.
The detailed network configurations are shown in Figure~\ref{fig: network-configurations}.

\subsection{Optical flow estimation}
\label{ssec: Optical flow estimation}

The optical flow estimation is mainly used to compute the warping matrix so that the information of the adjacent frames can be used for better latent HR frame restoration.
In this work, we use the PWC-Net~\cite{PWCNet} as our optical flow estimation model because it is effective in some video restoration tasks (e.g., video deblurring~\cite{CDVD/deblur/cvpr20}).
As the latent HR sequence $\{\mathrm{x}_i\}$ is not available, we compute the optical flow from the LR input sequence by:
\begin{equation}
	\mathrm{u}_{j\to i}  = \mathcal{N}_f(\mathrm{y}_j, \mathrm{y}_i), j=i-N,..., i-1, i+1,...,i+N,
	\label{eq: optical-flow}
\end{equation}
where $\mathcal{N}_f$ denotes the optical flow estimation network, which adopts the default network configurations of~\cite{PWCNet}.
With the estimated optical flow $\mathrm{u}_{j\to i}$, the warping operation, i.e., $\mathrm{F}_{j\to i}\mathrm{y}_j$, can be computed by applying the bilinear interpolation to $\mathrm{y}_j$.
Similar to the warping operation in~\cite{PWCNet}, we perform the warping operation on the features of LR input frames, where the features are extracted by the feature extraction network $\mathcal{N}_e$ as shown in Figure~\ref{fig: flow-chart}.
The detailed network configurations of $\mathcal{N}_e$ are shown in Figure~\ref{fig: network-configurations}.

\subsection{Latent HR frame restoration}
\label{ssec: Latent HR frame restoration}

Given the warped features of LR input frames, we estimate the latent frame $\mathrm{x}_i$ by:
\begin{equation}
	\mathrm{x}_i = \mathcal{N}_I(\mathcal{C}[\mathrm{y}_{i-N}^{e,w},..., \mathrm{y}_{i-1}^{e,w}, \mathrm{y}_i^e, \mathrm{y}_{i+1}^{e,w}, ..., \mathrm{y}_{i+N}^{e,w}]),
	\label{eq: hr-restoration}
\end{equation}
where $\mathcal{N}_I$ denotes the latent HR frame restoration network;
$\mathrm{y}_i^e$ denotes the extracted feature of the $i$-th LR frame (i.e., $\mathrm{y}_i^e = \mathcal{N}_e(\mathrm{y}_i)$)
and $\{\mathrm{y}_j^{e,w}\}_{j=i-N \& j\neq i}^{i+N}$ denotes the warped features according to the estimated optical flow (i.e., $\mathrm{y}_j^{e,w} = \mathrm{F}_{j\to i} \mathcal{N}_e(\mathrm{y}_j)$).
For the network architecture of $\mathcal{N}_I$, we adopt 20 ResBlocks~\cite{edsr} for feature reconstruction and use the pixel shuffle modules~\cite{shi2016real} for upsampling.
The detailed network configurations of $\mathcal{N}_I$ are shown in Figure~\ref{fig: network-configurations}.

\subsection{Self-supervised learning}
\label{ssec: Self-supervised learning}

As the ground truth HR videos and blur kernels are not available, a straightforward way to train the networks $\mathcal{N}_k$, $\mathcal{N}_f$, $\mathcal{N}_e$ and $\mathcal{N}_I$ is to minimize the following loss function:
\begin{equation}
	\mathcal{L}_{self} = \rho(\mathbf{SK}_i\mathrm{x}_i - \mathrm{y}_i),
	\label{eq: self-supervised-loss-lr}
\end{equation}
where $\rho(\cdot)$ denotes a robust function which is usually taken $L_1$-norm or $L_2$-norm.
However, directly minimizing~\eqref{eq: self-supervised-loss-lr} usually leads to trivial solutions.
To overcome this problem, we explore the properties of the blur kernels and the image formation model~\eqref{eq: sr-formation-convolution} to constrain the blur kernel estimation and latent HR frame restoration process.

As the elements of the blur kernels are usually sparse, we develop a hyper-Laplacian prior to model the sparse property of the output of the network $\mathcal{N}_k$:
\begin{equation}
	\mathcal{L}_{k} = \|K_i\|^{\alpha},
	\label{eq: sparse-kernel-loss}
\end{equation}
where $\alpha$ denotes a hyperparameter whose value is usually taken $0.5$ according to~\cite{kernelgan}.

To regularize the latent HR frame restoration process, we develop a video degradation constraint based on the image formation model~\eqref{eq: sr-formation-convolution}.
Before presenting the video degradation constraint, we first introduce the following property.

{\flushleft \bf Property}: \emph{
	Let $\mathbf{K}_i^*$, $\mathbf{F}_{i\to j}^*$, and $\mathcal{F}$ denote the ground truth blur kernel matrix, warping matrix, and the exact LR-to-HR mapping function.
	That is, the following equation
	\begin{equation}\label{eq: inverse-mapping-1}
		\mathrm{x}_i = \mathcal{F}(\mathrm{y}_{i-N};...; \mathrm{y}_{i-1}; \mathrm{y}_{i}; \mathrm{y}_{i+1};...; \mathrm{y}_{i+N})
	\end{equation}
	strictly holds.
	Therefore, for any videos $\{\mathrm{L}_j\}$, if $\mathrm{L}_j =\mathbf{S}\mathbf{K}_j^*\mathbf{F}_{i\to j}^*\mathrm{H}_i$, we have:
	\begin{equation}\label{eq: inverse-mapping-2}
		\mathrm{H}_i = \mathcal{F}(\mathrm{L}_{i-N};...; \mathrm{L}_{i-1}; \mathrm{L}_{i};\mathrm{L}_{i+1};...; \mathrm{L}_{i+N}).
	\end{equation}
}

We note that we can generate auxiliary LR frames $\{\mathrm{L}_i\}$ by applying the estimated blur kernel and the optical flow to any HR sequence $\{\mathrm{H}_i\}$ according to the image formation model~\eqref{eq: sr-formation-convolution}.
Instead of using additional HR reference videos, we directly use $\{\mathrm{y}_i\}$ as the HR sequence to generate auxiliary LR frames $\{\mathrm{L}_i\}$.
Based on the above property, if the latent HR frame restoration network $\mathcal{N}_I$ is accurately estimated, the output of the network $\mathcal{N}_I$ should be close to $\mathrm{y}_i$.
Thus, we develop a constraint to regularize the network $\mathcal{N}_I$ by:
\begin{equation}
	\mathcal{L}_{I} = \rho(\mathcal{N}_I(\mathcal{C}[\mathrm{L}_{i-N}^{e,w},..., \mathrm{L}_{i-1}^{e,w}, \mathrm{L}_i^e, \mathrm{L}_{i+1}^{e,w}, ..., \mathrm{L}_{i+N}^{e,w}]) - \mathrm{y}_i),
	\label{eq: self-image-loss}
\end{equation}
where $\mathrm{L}_i^e$ denotes the extracted feature of the auxiliary LR frame $\mathrm{L}_i$;
$\{\mathrm{L}_j^{e,w}\}_{j=i-N \& j\neq i}^{i+N}$ denotes the warped features according to the estimated optical flow $\mathcal{N}_f(\mathrm{L}_j, \mathrm{L}_i)$ and $\{\mathrm{L}_i\} = \{\mathbf{SK}_i\mathrm{y}_i\}$.

Based on the above considerations, the proposed self-supervised learning for the video SR can be achieved by minimizing:
\begin{equation}
	\mathcal{L} = \mathcal{L}_{self} + \lambda \mathcal{L}_{I} + \gamma\mathcal{L}_{k},
	\label{eq: whole-loss}
\end{equation}
where $\lambda$ and $\gamma$ are weight parameters.
We will demonstrate the  effectiveness of self-supervised learning in Section~\ref{ssec: analysis-self-supervised-learning}.

\subsection{Implementation details and datasets}

\subsubsection{Implementation details}

We train the networks $\mathcal{N}_k$, $\mathcal{N}_e$ and $\mathcal{N}_I$ from scratch, where the learning rate for these three networks is initialized to be $10^{-4}$.
We use the pre-trained model by~\cite{PWCNet} as the initialization of the optical flow estimation network $\mathcal{N}_f$. The learning rate for $\mathcal{N}_f$ is initialized to be $10^{-6}$ as it adopts the pre-trained model.
The ADAM optimizer~\cite{adam} with parameters $\beta_1 = 0.9$, $\beta_2 = 0.999$, and $\epsilon = 10^{-8}$ is used for the network training.
All the networks are jointly trained in an end-to-end manner based on the self-supervised learning method.
The learning rate values decrease to half after every 100 epochs, and 200 epochs are used.
For the weight parameters in the self-supervised loss function~\eqref{eq: whole-loss}, we empirically set $\lambda$ and $\gamma$ to be $1$ and $0.04$.
Similar to~\cite{kernelgan}, we further use the boundary loss and the center loss to constrain the estimated blur kernels by encouraging the boundary values to be zero and the center of mass to be at the center of the kernel.
The training code, models and experimental results used in the paper will be available at \url{https://github.com/csbhr/Self-Blind-VSR}.

\subsubsection{Datasets}

We first evaluate the proposed algorithm on the synthetic datasets, which are generated with the different Gaussian blur kernels and  the realistic motion blur kernels~\cite{kernelgan} based on the image formation model~\eqref{eq: sr-formation-convolution}.
Then, we further perform the qualitative evaluation on the real-world videos to evaluate the generalization ability of the proposed method.
As the proposed method focuses on the video SR problem where the ground truth HR videos of LR inputs are unavailable and the degradation parameters in~\eqref{eq: sr-formation-convolution} are unknown, the ground truth HR videos in synthetic datasets are only used for metrics calculation, not for deep model training.

To generate the synthetic dataset for quantitative evaluation, the REDS dataset~\cite{reds} is used as the training dataset, and the commonly used video SR datasets the REDS4 dataset (split from~\cite{reds} by~\cite{edvr}), the VID4 dataset~\cite{Bayesian/vsr/pami/LiuS14}, and the SPMCS dataset~\cite{taoxin/iccv17} are adopted as benchmarks to evaluate our method.
When generating the LR videos, we apply the blurring operation with the downsampling operation to each HR video according to the image formation model~\eqref{eq: sr-formation-convolution}.
For the blurring operation, we use the different Gaussian blur kernels and the realistic motion blur kernels from~\cite{kernelgan}.
For the Gaussian blur kernels, the standard variation values range from $0.4$ to $2$.
The downsampling operation is used to extract the pixels based on the scale factor, and the scale factor is set to be $4$.
%

%-------------------------------------------------------
\section{Experimental Results}
\label{sec: experimental-results}
%----------------------------
\begin{table*}[!t]
	\caption{Comparisons of the video SR results by the state-of-the-art methods on the REDS4 dataset~\cite{edvr} with different Gaussian blur kernels in terms of PSNR and SSIM. The proposed method generates the results with the highest values.}
	\label{tab:gaussian-reds4-quantitative}
	\footnotesize
	\resizebox{0.99\textwidth}{!}{
		\centering
		\begin{tabular}{lccccccccccc}
			\toprule
			Methods  &Bicubic  &RCAN~\cite{RCAN}  &MZSR~\cite{MZSR}  &ZSSR~\cite{zssr}  &KernelGAN~\cite{kernelgan}  &IKC~\cite{IKC}  &RBPN~\cite{VDBPN/cvpr19}  &DUF~\cite{DUF}   &TDAN~\cite{TDAN/CVPR2020}  &EDVR~\cite{edvr}  &Ours  \\
			\hline
			Supervised  &\XSolidBrush  &\CheckmarkBold  &\XSolidBrush  &\XSolidBrush  &\XSolidBrush  &\CheckmarkBold  &\CheckmarkBold  &\CheckmarkBold  &\CheckmarkBold  &\CheckmarkBold  &\XSolidBrush  \\
			Blind  &-  &\XSolidBrush  &\XSolidBrush  &\XSolidBrush  &\CheckmarkBold  &\CheckmarkBold  &\XSolidBrush  &\XSolidBrush  &\XSolidBrush  &\XSolidBrush  &\CheckmarkBold  \\
			\hline
			PSNR  &25.99  &27.39  &27.04  &26.67  &19.10  &27.08  &27.00  &26.82  &27.64  &27.91  &\bf{29.23}  \\
			SSIM  &0.7297  &0.7972  &0.7727  &0.7623  &0.5403  &0.7940  &0.8054  &0.8149  &0.8050  &0.8336  &\bf{0.8453}  \\
			\bottomrule
		\end{tabular}
	}
\end{table*}
%----------------------------

%----------------------------
\begin{figure*}[!t]\footnotesize
	\begin{center}
		\begin{tabular}{ccccccc}
			\multicolumn{3}{c}{\multirow{5}*[51.8pt]{\includegraphics[width=0.368\linewidth, height = 0.247\linewidth]{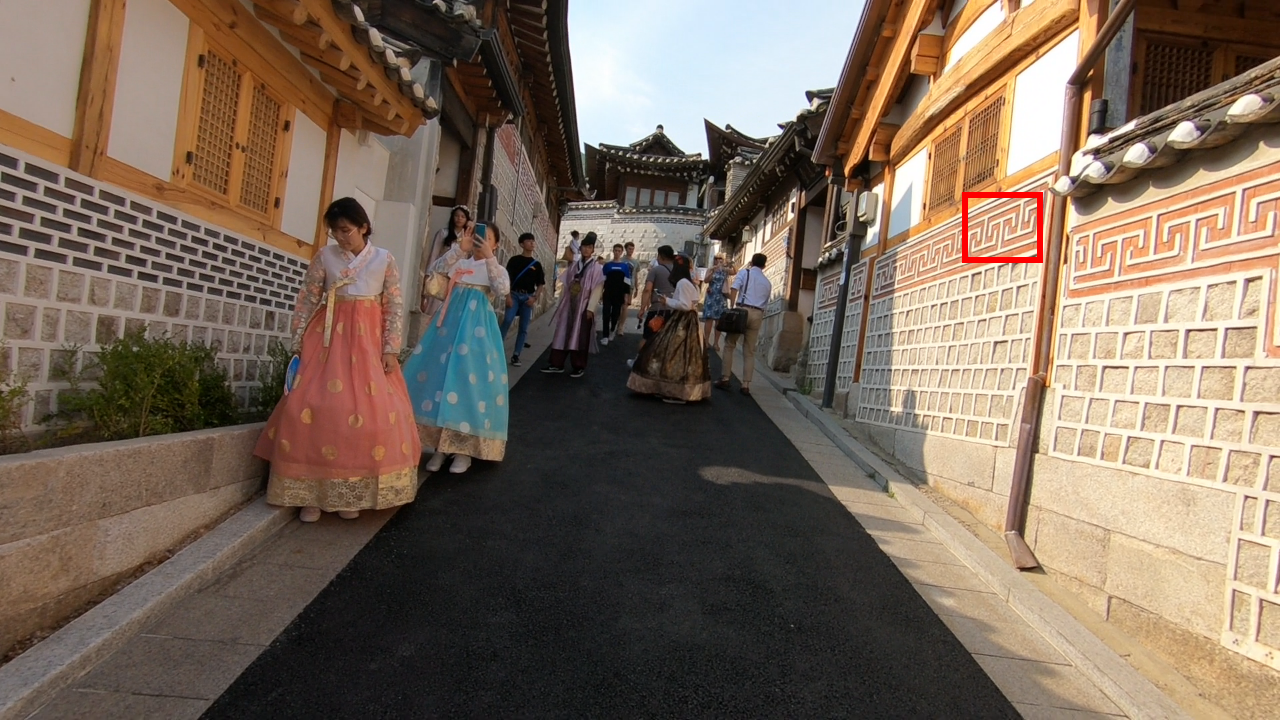}}}&\hspace{-4.5mm}
			\includegraphics[width=0.15\linewidth, height = 0.112\linewidth]{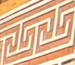} &\hspace{-4.5mm}
			\includegraphics[width=0.15\linewidth, height = 0.112\linewidth]{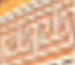} &\hspace{-4.5mm}
			\includegraphics[width=0.15\linewidth, height = 0.112\linewidth]{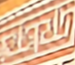} &\hspace{-4.5mm}
			\includegraphics[width=0.15\linewidth, height = 0.112\linewidth]{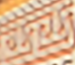} \\
			\multicolumn{3}{c}{~} &\hspace{-4.5mm}  (b) HR patch &\hspace{-4.5mm}  (c) Bicubic &\hspace{-4.5mm}  (d) IKC~\cite{IKC}  &\hspace{-4.5mm}  (e) ZSSR~\cite{zssr}\\
			\multicolumn{3}{c}{~} & \hspace{-4.5mm}
			\includegraphics[width=0.15\linewidth, height = 0.112\linewidth]{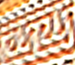} & \hspace{-4.5mm}
			\includegraphics[width=0.15\linewidth, height = 0.112\linewidth]{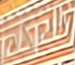} & \hspace{-4.5mm}
			\includegraphics[width=0.15\linewidth, height = 0.112\linewidth]{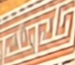} & \hspace{-4.5mm}
			\includegraphics[width=0.15\linewidth, height = 0.112\linewidth]{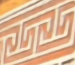} \\
			\multicolumn{3}{c}{\hspace{-4.5mm} (a) Ground truth HR frame} &  \hspace{-4.5mm} (f) KernelGAN~\cite{kernelgan} &\hspace{-4.5mm}  (g) RBPN~\cite{VDBPN/cvpr19} &\hspace{-4.5mm}  (h) EDVR~\cite{edvr} & \hspace{-4.5mm} (i) Ours\\
		\end{tabular}
	\end{center}
	\caption{Comparison of the video SR results on the REDS4 dataset~\cite{edvr} ($\times 4$). Our method recovers high-quality frame with clearer structures.}
	\label{fig: results-reds}
\end{figure*}
%----------------------------

%----------------------------
\begin{table*}[!t]
	\caption{Comparisons of the video SR results by the state-of-the-art methods on the VID4 dataset~\cite{Bayesian/vsr/pami/LiuS14} with different Gaussian blur kernels in terms of PSNR and SSIM. The proposed method generates the results with the highest values.}
	\label{tab:gaussian-vid4-quantitative}
	\footnotesize
	\resizebox{0.99\textwidth}{!}{
		\centering
		\begin{tabular}{lccccccccccc}
			\toprule
			Methods  &Bicubic  &RCAN~\cite{RCAN}  &MZSR~\cite{MZSR}  &ZSSR~\cite{zssr}  &KernelGAN~\cite{kernelgan}  &IKC~\cite{IKC}  &RBPN~\cite{VDBPN/cvpr19}  &DUF~\cite{DUF}  &TDAN~\cite{TDAN/CVPR2020}  &EDVR~\cite{edvr}  &Ours  \\
			\hline
			PSNR  &22.27  &23.13  &22.99  &22.83  &16.07  &22.80  &22.07  &23.04  &23.94  &23.62  &\bf{24.59}  \\
			SSIM  &0.6159  &0.7008  &0.6699  &0.6655  &0.4403  &0.6955  &0.6917  &0.7622  &0.7457  &0.7506  &\bf{0.7629}  \\
			\bottomrule
		\end{tabular}
	}
\end{table*}
%----------------------------

%----------------------------
\begin{figure*}[!t]\footnotesize
	\begin{center}
		\begin{tabular}{ccccccc}
			\multicolumn{3}{c}{\multirow{5}*[61.2pt]{\includegraphics[width=0.368\linewidth, height = 0.284\linewidth]{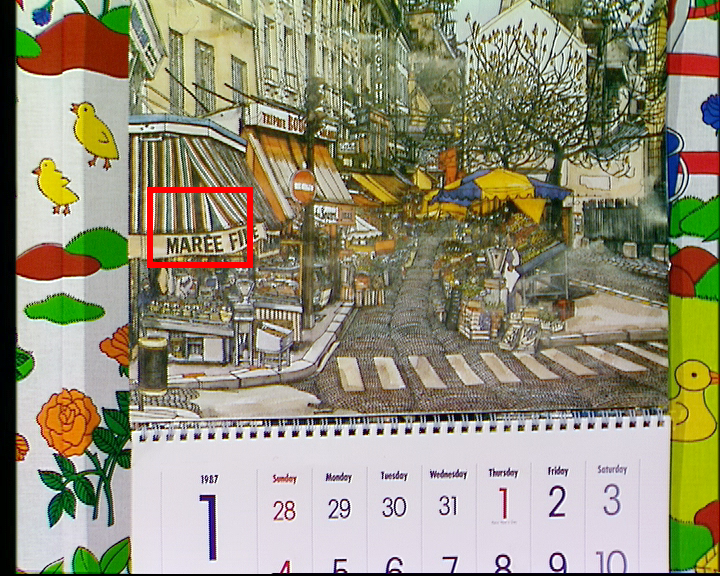}}}&\hspace{-4.5mm}
			\includegraphics[width=0.15\linewidth, height = 0.13\linewidth]{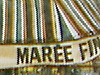} &\hspace{-4.5mm}
			\includegraphics[width=0.15\linewidth, height = 0.13\linewidth]{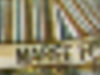} &\hspace{-4.5mm}
			\includegraphics[width=0.15\linewidth, height = 0.13\linewidth]{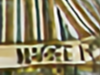} &\hspace{-4.5mm}
			\includegraphics[width=0.15\linewidth, height = 0.13\linewidth]{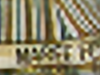} \\
			\multicolumn{3}{c}{~} &\hspace{-4.5mm}  (b) HR patch &\hspace{-4.5mm}  (c) Bicubic &\hspace{-4.5mm}  (d) IKC~\cite{IKC}  &\hspace{-4.5mm}  (e) ZSSR~\cite{zssr}\\
			\multicolumn{3}{c}{~} & \hspace{-4.5mm}
			\includegraphics[width=0.15\linewidth, height = 0.13\linewidth]{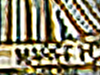} & \hspace{-4.5mm}
			\includegraphics[width=0.15\linewidth, height = 0.13\linewidth]{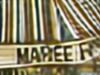} & \hspace{-4.5mm}
			\includegraphics[width=0.15\linewidth, height = 0.13\linewidth]{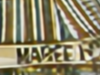} & \hspace{-4.5mm}
			\includegraphics[width=0.15\linewidth, height = 0.13\linewidth]{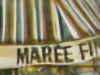} \\
			\multicolumn{3}{c}{\hspace{-4.5mm} (a) Ground truth HR frame} &  \hspace{-4.5mm} (f) KernelGAN~\cite{kernelgan} &\hspace{-4.5mm}  (g) RBPN~\cite{VDBPN/cvpr19} &\hspace{-4.5mm}  (h) EDVR~\cite{edvr} & \hspace{-4.5mm} (i) Ours\\
		\end{tabular}
	\end{center}
	\caption{Comparison of the video SR results on the VID4 dataset~\cite{Bayesian/vsr/pami/LiuS14} ($\times 4$). Our method recovers the frame with clearer structures and characters.}
	\label{fig: results-vid4}
\end{figure*}
%----------------------------

%----------------------------
\begin{table*}[!t]
	\caption{Comparisons of the video SR results by the state-of-the-art methods on the SPMCS dataset~\cite{taoxin/iccv17} with different Gaussian blur kernels in terms of PSNR and SSIM. The proposed method generates the results with the highest values.}
	\label{tab:gaussian-spmcs-quantitative}
	\footnotesize
	\resizebox{0.99\textwidth}{!}{
		\centering
		\begin{tabular}{lccccccccccc}
			\toprule
			Methods  &Bicubic  &RCAN~\cite{RCAN}  &MZSR~\cite{MZSR}  &ZSSR~\cite{zssr}  &KernelGAN~\cite{kernelgan}  &IKC~\cite{IKC}  &RBPN~\cite{VDBPN/cvpr19}  &DUF~\cite{DUF}  &TDAN~\cite{TDAN/CVPR2020}  &EDVR~\cite{edvr}  &Ours  \\
			\hline
			PSNR  &25.51  &27.17  &26.43  &26.15  &18.90  &26.71  &26.16  &26.50  &27.51  &27.23  &\bf{27.77}  \\
			SSIM  &0.7241  &0.8056  &0.7706  &0.7606  &0.5501  &0.7920  &0.7941  &0.8129  &0.8149  &0.8055  &\bf{0.8184}  \\
			\bottomrule
		\end{tabular}
	}
\end{table*}
%----------------------------

%----------------------------
\begin{figure*}[!t]\footnotesize
	\begin{center}
		\begin{tabular}{ccccccc}
			\multicolumn{3}{c}{\multirow{5}*[61.5pt]{\includegraphics[width=0.368\linewidth, height = 0.285\linewidth]{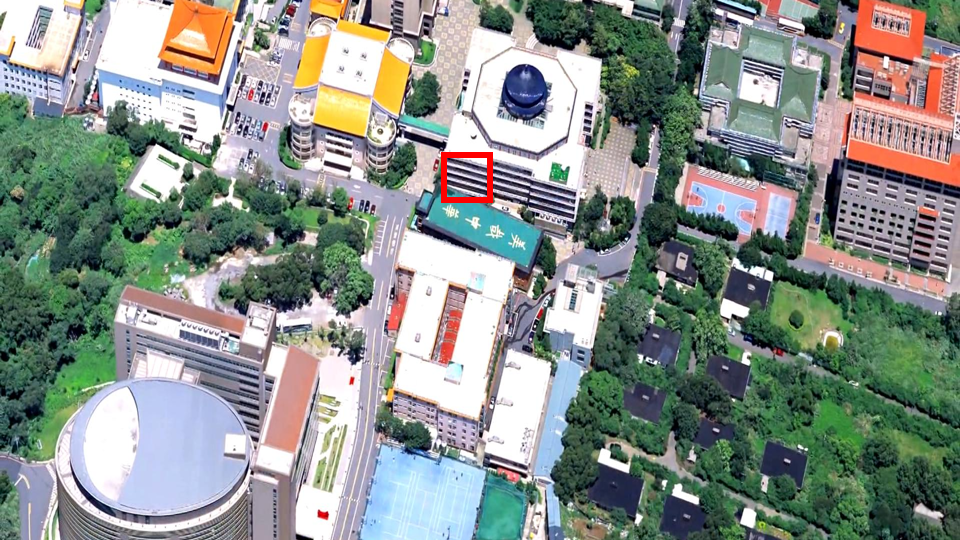}}}&\hspace{-4.5mm}
			\includegraphics[width=0.15\linewidth, height = 0.13\linewidth]{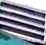} &\hspace{-4.5mm}
			\includegraphics[width=0.15\linewidth, height = 0.13\linewidth]{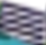} &\hspace{-4.5mm}
			\includegraphics[width=0.15\linewidth, height = 0.13\linewidth]{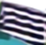} &\hspace{-4.5mm}
			\includegraphics[width=0.15\linewidth, height = 0.13\linewidth]{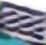} \\
			\multicolumn{3}{c}{~} &\hspace{-4.5mm}  (b) HR patch &\hspace{-4.5mm}  (c) Bicubic &\hspace{-4.5mm}  (d) IKC~\cite{IKC}  &\hspace{-4.5mm}  (e) MZSR~\cite{MZSR}\\
			\multicolumn{3}{c}{~} & \hspace{-4.5mm}
			\includegraphics[width=0.15\linewidth, height = 0.13\linewidth]{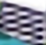} & \hspace{-4.5mm}
			\includegraphics[width=0.15\linewidth, height = 0.13\linewidth]{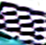} & \hspace{-4.5mm}
			\includegraphics[width=0.15\linewidth, height = 0.13\linewidth]{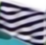} & \hspace{-4.5mm}
			\includegraphics[width=0.15\linewidth, height = 0.13\linewidth]{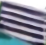} \\
			\multicolumn{3}{c}{\hspace{-4.5mm} (a) Ground truth HR frame} &  \hspace{-4.5mm} (f) ZSSR~\cite{zssr} &\hspace{-4.5mm}  (g) KernelGAN~\cite{kernelgan} &\hspace{-4.5mm}  (h) EDVR~\cite{edvr} & \hspace{-4.5mm} (i) Ours\\
		\end{tabular}
	\end{center}
	\caption{Comparison of the video SR results on the SPMCS dataset~\cite{taoxin/iccv17} ($\times 4$). Our method recovers the frame with correct structures.}
	\label{fig: results-spmcs}
\end{figure*}
%----------------------------

%----------------------------
\begin{table*}[!t]
	\caption{Comparisons of the video SR results by the state-of-the-art methods on the REDS4 dataset~\cite{edvr} with realistic motion blur kernels~\cite{kernelgan} in terms of PSNR and SSIM. The proposed method generates the results with the highest values.}
	\label{tab:motion-reds4-quantitative}
	\footnotesize
	\resizebox{0.99\textwidth}{!}{
		\centering
		\begin{tabular}{lccccccccccc}
			\toprule
			Methods  &Bicubic  &RCAN~\cite{RCAN}  &MZSR~\cite{MZSR}  &ZSSR~\cite{zssr}  &KernelGAN~\cite{kernelgan}  &IKC~\cite{IKC}  &RBPN~\cite{VDBPN/cvpr19}  &DUF~\cite{DUF}  &TDAN~\cite{TDAN/CVPR2020}  &EDVR~\cite{edvr}  &Ours  \\
			\hline
			PSNR  &25.78  &26.84  &26.87  &26.38  &26.35  &26.64  &26.83  &26.51  &27.13  &27.27  &\bf{28.42}  \\
			SSIM  &0.7197  &0.7797  &0.7639  &0.7503  &0.7507  &0.7783  &0.7957  &0.7972  &0.7879  &0.8084  &\bf{0.8172}  \\
			\bottomrule
		\end{tabular}
	}
\end{table*}
%----------------------------

%----------------------------
\begin{figure*}[!t]\footnotesize
	\begin{center}
		\begin{tabular}{ccccccc}
			\multicolumn{3}{c}{\multirow{5}*[61.5pt]{\includegraphics[width=0.368\linewidth, height = 0.285\linewidth]{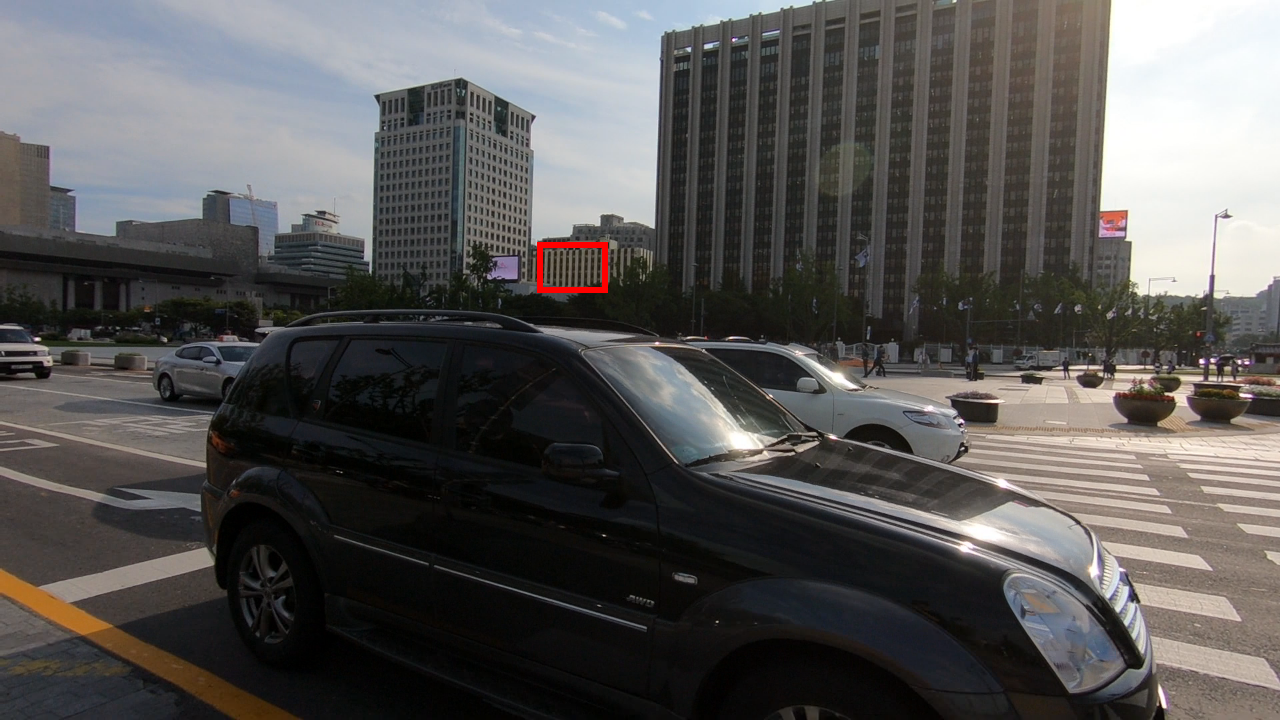}}}&\hspace{-4.5mm}
			\includegraphics[width=0.15\linewidth, height = 0.13\linewidth]{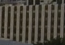} &\hspace{-4.5mm}
			\includegraphics[width=0.15\linewidth, height = 0.13\linewidth]{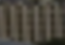} &\hspace{-4.5mm}
			\includegraphics[width=0.15\linewidth, height = 0.13\linewidth]{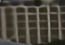} &\hspace{-4.5mm}
			\includegraphics[width=0.15\linewidth, height = 0.13\linewidth]{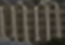} \\
			\multicolumn{3}{c}{~} &\hspace{-4.5mm}  (b) HR patch &\hspace{-4.5mm}  (c) Bicubic &\hspace{-4.5mm}  (d) IKC~\cite{IKC}  &\hspace{-4.5mm}  (e) ZSSR~\cite{zssr}\\
			\multicolumn{3}{c}{~} & \hspace{-4.5mm}
			\includegraphics[width=0.15\linewidth, height = 0.13\linewidth]{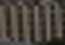} & \hspace{-4.5mm}
			\includegraphics[width=0.15\linewidth, height = 0.13\linewidth]{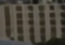} & \hspace{-4.5mm}
			\includegraphics[width=0.15\linewidth, height = 0.13\linewidth]{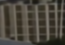} & \hspace{-4.5mm}
			\includegraphics[width=0.15\linewidth, height = 0.13\linewidth]{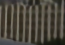} \\
			\multicolumn{3}{c}{\hspace{-4.5mm} (a) Ground truth HR frame} &  \hspace{-4.5mm} (f) KernelGAN~\cite{kernelgan} &\hspace{-4.5mm}  (g) RBPN~\cite{VDBPN/cvpr19} &\hspace{-4.5mm}  (h) EDVR~\cite{edvr} & \hspace{-4.5mm} (i) Ours\\
		\end{tabular}
	\end{center}
	\caption{Comparison of the video SR results on the REDS4 dataset~\cite{edvr} with realistic motion blur kernels ($\times 4$). Our method recovers the frame with more correct structures.}
	\label{fig: motion-results-reds4}
\end{figure*}
%----------------------------

%----------------------------
\begin{figure*}[!t]\footnotesize
	\begin{center}
		\begin{tabular}{ccccccc}
			\multicolumn{3}{c}{\multirow{5}*[66.5pt]{\includegraphics[width=0.368\linewidth, height = 0.305\linewidth]{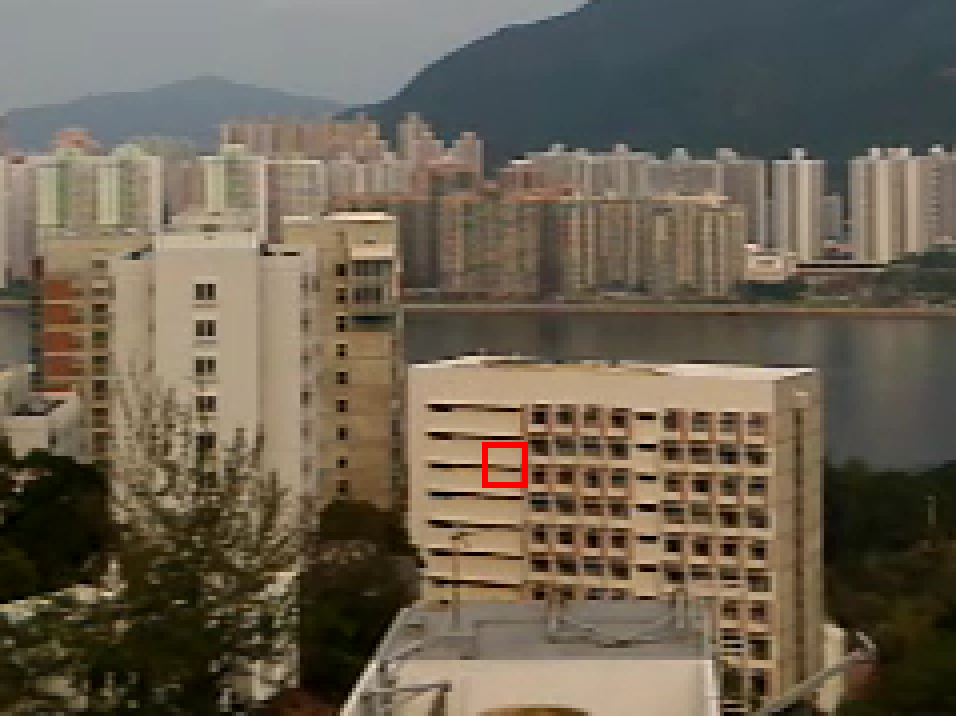}}}&\hspace{-4.5mm}
			\includegraphics[width=0.15\linewidth, height = 0.14\linewidth]{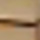} &\hspace{-4.5mm}
			\includegraphics[width=0.15\linewidth, height = 0.14\linewidth]{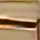} &\hspace{-4.5mm}
			\includegraphics[width=0.15\linewidth, height = 0.14\linewidth]{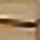} &\hspace{-4.5mm}
			\includegraphics[width=0.15\linewidth, height = 0.14\linewidth]{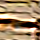} \\
			\multicolumn{3}{c}{~} &\hspace{-4.5mm}  (b) Bicubic &\hspace{-4.5mm}  (c) IKC~\cite{IKC} &\hspace{-4.5mm}  (d) ZSSR~\cite{zssr}  &\hspace{-4.5mm}  (e) KernelGAN~\cite{kernelgan}\\
			\multicolumn{3}{c}{~} & \hspace{-4.5mm}
			\includegraphics[width=0.15\linewidth, height = 0.14\linewidth]{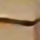} & \hspace{-4.5mm}
			\includegraphics[width=0.15\linewidth, height = 0.14\linewidth]{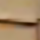} & \hspace{-4.5mm}
			\includegraphics[width=0.15\linewidth, height = 0.14\linewidth]{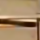} & \hspace{-4.5mm}
			\includegraphics[width=0.15\linewidth, height = 0.14\linewidth]{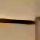} \\
			\multicolumn{3}{c}{\hspace{-4.5mm} (a) LR input frame} &  \hspace{-4.5mm} (f)  MZSR~\cite{MZSR} &\hspace{-4.5mm}  (g) RBPN~\cite{VDBPN/cvpr19} &\hspace{-4.5mm}  (h) EDVR~\cite{edvr} & \hspace{-4.5mm} (i) Ours\\
		\end{tabular}
	\end{center}
	\caption{Super-resolving real-world videos ($\times 4$). The proposed method recovers the frame with fewer artifacts and clearer structures.}
	\label{fig: results-real}
\end{figure*}
%----------------------------

As our method aims to solve the blind video SR problem in a self-supervised manner, few methods have been proposed for this problem.
To evaluate the performance of our algorithm, we still compare it against the related self-supervised single image SR methods (ZSSR~\cite{zssr}, KernelGAN~\cite{kernelgan}, MZSR~\cite{MZSR}), the supervised single image SR method (RCAN~\cite{RCAN}), the blind single image SR method (IKC~\cite{IKC}), and state-of-the-art video SR methods based on supervised learning (RBPN~\cite{VDBPN/cvpr19}, DUF~\cite{DUF}, TDAN~\cite{TDAN/CVPR2020}, EDVR~\cite{edvr}).
For ZSSR~\cite{zssr}, the default kernels are Bicubic ones for evaluations.
For KernelGAN~\cite{kernelgan}, we use the released codes with the default settings, where ZSSR~\cite{zssr} is used to restore HR images for evaluations.

For the comparisons with the supervised learning-based methods, as the ground truth HR videos are not used for training, the supervised SR methods (RCAN~\cite{RCAN}, IKC~\cite{IKC}, RBPN~\cite{VDBPN/cvpr19}, DUF~\cite{DUF}, TDAN~\cite{TDAN/CVPR2020}, EDVR~\cite{edvr}) are evaluated with the officially provided pre-trained models for fairness.
For the synthetic videos, we use the PSNR and SSIM as the quantitative evaluation metrics.

\subsection{Quantitative evaluations}

Table~\ref{tab:gaussian-reds4-quantitative}-~\ref{tab:gaussian-spmcs-quantitative} show the quantitative evaluations on the benchmark datasets with different Gaussian blur kernels, where our method generates the results with the highest PSNR and SSIM values than state-of-the-art methods.
We note that the RCAN~\cite{RCAN} method does not generate high-quality SR results as it assumes that the blur kernel is known.
Although the blind image SR methods~\cite{IKC,kernelgan} involve the blur kernel estimation, they are designed for single images and do not perform well on the video SR problems. It shows that the PSNR values of our method are at least 1.06dB higher than these blind image SR methods.
As the supervised video SR methods~\cite{VDBPN/cvpr19,DUF,TDAN/CVPR2020,edvr} assume that the blur kernel is known (e.g., Bicubic) and train the deep models on the datasets with the predefined blur kernels, they cannot be generalized well on the datasets with unknown blur kernels.
In contrast, our method explicitly involves the blur kernel estimation and solves the deep models in a self-supervised manner. Thus, it can generate favorable results against these video SR methods without any paired or unpaired training datasets.

We further evaluate our method when the blur kernels are complex realistic ones~\cite{kernelgan} in Table~\ref{tab:motion-reds4-quantitative}.
We note that the methods~\cite{zssr,MZSR,kernelgan} based on the self-supervised learning algorithm, the supervised image SR~\cite{RCAN,IKC} and the supervised video SR methods~\cite{VDBPN/cvpr19,DUF,TDAN/CVPR2020,edvr} do not generate good results.
In contrast, our method generates the results with higher PSNR and SSIM values, which demonstrates that our method generalizes well compared to the existing methods.

\subsection{Qualitative evaluations}

Figure~\ref{fig: results-reds} shows some visual comparisons of SR results generated by the evaluated methods on the REDS4 dataset~\cite{reds} with a scale factor of 4.
We note that the ZSSR method~\cite{zssr} does not generate a clearer image as it does not involve the blur kernel estimation (Figure~\ref{fig: results-reds}(e)).
Although the KernelGAN method~\cite{kernelgan} explicitly estimates blur kernels from LR images and can use the ZSSR method to super-resolve images, the generated results contain significant artifacts due to the imperfect blur kernels (Figure~\ref{fig: results-reds}(f)).
To correct errors of the blur kernels, the IKC method~\cite{IKC} develops an effective iterative kernel correction method. Although the quality of the SR results is improved (Figure~\ref{fig: results-reds}(d)), the structural details are still not restored well as this method is not designed for the video SR problem.
Although the RBPN method~\cite{VDBPN/cvpr19} and EDVR method~\cite{edvr} are developed to solve video SR, these methods assume that the blur kernel is known and do not solve the blind video SR problem well as shown in Figure~\ref{fig: results-reds}(g) and (h).
In contrast, although our method does not require the HR videos as the supervision, it generates much clearer frames with better structural details as shown in Figure~\ref{fig: results-reds}(i). This further demonstrates the effectiveness of the proposed method on the blind video SR problem.

Figure~\ref{fig: results-vid4} and Figure~\ref{fig: results-spmcs} show some examples from the VID4 dataset~\cite{Bayesian/vsr/pami/LiuS14} and the SPMCS dataset~\cite{taoxin/iccv17}.
Our method generates the frames with finer details, where the characters in the restored frames are recognizable.
In addition, Figure~\ref{fig: motion-results-reds4} shows the visual comparisons on the REDS4 dataset~\cite{edvr} with the realistic motion blur kernels from~\cite{kernelgan}. It shows that our method also can recover more correct structures than these state-of-the-art methods.

We further evaluate our method on real-world videos. Figure~\ref{fig: results-real} shows comparisons on a real-world LR video.
We note that state-of-the-art methods do not restore the structural details well. The restored results are still blurry or contain significant artifacts as shown in Figure~\ref{fig: results-real}(c)-(h).
In contrast, our algorithm generates a much clearer frame.
%

%-------------------------------------------------------
\section{Analysis and Discussions}
\label{sec: analysis}

In this section, we provide further analysis of the proposed self-supervised learning method and discuss the major differences from the closely-related methods.

%----------------------------
\begin{table*}[!t]
	\caption{Effectiveness of the proposed self-supervised learning method on video SR.
	}
	\label{tab: unsupervised-comparisons}
	\footnotesize
	\resizebox{0.99\textwidth}{!}{
		\centering
		\begin{tabular}{lccccc}
			\toprule
			Methods~~~~~~~~~~~~  &~~~~~~w/o~\eqref{eq: self-image-loss} \&~\eqref{eq: sparse-kernel-loss}~~~~~~  &~~~~~~~~w/o~\eqref{eq: sparse-kernel-loss}~~~~~~~~  &~~~~~~~~~w/o~\eqref{eq: self-image-loss}~~~~~~~~~  &~~~~w/o detach in~\eqref{eq: self-supervised-loss-lr} ~~~~  & ~~~~~~~~~~Ours~~~~~~~~~~\\
			\hline
			REDS4 dataset  &25.73/0.7187  &27.66/0.7975  &26.12/0.7401  &26.71/0.7632  &\bf{29.23/0.8453}\\
			VID4 dataset  &22.04/0.6015  &23.49/0.7039  &22.44/0.6355  &23.03/0.6813  &\bf{24.59/0.7629}\\
			SPMCS dataset  &25.23/0.7114  &26.60/0.7766  &25.67/0.7373  &25.86/0.7514  &\bf{27.77/0.8184}\\
			\bottomrule
		\end{tabular}
	}
\end{table*}
%----------------------------

%-------------------------------
\begin{figure}[!t]\footnotesize
	\centering
	\begin{tabular}{ccc}
		\includegraphics[width=0.32\linewidth]{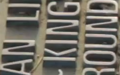} &\hspace{-4mm}
		\includegraphics[width=0.32\linewidth]{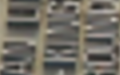} &\hspace{-4mm}
		\includegraphics[width=0.32\linewidth]{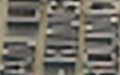}  \\
		(a) HR patch &\hspace{-4mm}  (b) Bicubic & \hspace{-4mm} (c)  ZSSR~\cite{zssr} \\
		\includegraphics[width=0.32\linewidth]{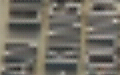} &\hspace{-4mm}
		\includegraphics[width=0.32\linewidth]{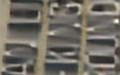} &\hspace{-4mm}
		\includegraphics[width=0.32\linewidth]{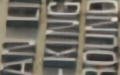}  \\
		(d) { w/o~\eqref{eq: sparse-kernel-loss} \&~\eqref{eq: self-image-loss}}  &\hspace{-4mm}  (e) { w/o~\eqref{eq: sparse-kernel-loss}} &\hspace{-4mm}  (f) Ours \\
	\end{tabular}
	\caption{Effectiveness of the self-supervised learning.}
	\label{fig: self-supervised-learning-analysis}
\end{figure}
%-------------------------------

\subsection{Effectiveness of the self-supervised learning}
\label{ssec: analysis-self-supervised-learning}

To demonstrate the effectiveness of the proposed self-supervised learning, we disable the constraint~\eqref{eq: self-image-loss} and~\eqref{eq: sparse-kernel-loss} in the proposed method for fair comparisons.
Table~\ref{tab: unsupervised-comparisons} shows the quantitative evaluations on the benchmark datasets.
We note that the method without the constraint~\eqref{eq: self-image-loss} and~\eqref{eq: sparse-kernel-loss} does not generate high-quality videos.
The PSNR values of this baseline method are at least 2.54dB lower than that of the proposed method, suggesting the effectiveness of the proposed self-supervised learning on the video SR problem.

\begin{figure}[!t]\footnotesize
	\centering
	\begin{tabular}{cccc}
		\includegraphics[width=0.25\linewidth]{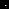} &\hspace{-4mm}
		\includegraphics[width=0.25\linewidth]{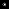} &\hspace{-2.5mm}
		\includegraphics[width=0.25\linewidth]{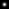}  \\
		(a) w/o~\eqref{eq: sparse-kernel-loss} \&~\eqref{eq: self-image-loss} &\hspace{-4mm}  (b) Ours & \hspace{-4mm} (c) GT\\
	\end{tabular}
	\caption{Blur kernel estimation results by baseline methods.}
	\label{fig: kernels}
\end{figure}

Figure~\ref{fig: self-supervised-learning-analysis}(d) shows the estimated SR results by the method without the constraint~\eqref{eq: self-image-loss} and~\eqref{eq: sparse-kernel-loss}, where there exist significant artifacts in the restored images.
In addition, the estimated blur kernel by the method without the constraint~\eqref{eq: self-image-loss} and~\eqref{eq: sparse-kernel-loss} looks like a delta kernel (see Figure~\ref{fig: kernels}(a)), which further verifies our claims in Section~\ref{ssec: Self-supervised learning}.

We further quantitatively evaluate the constraint~\eqref{eq: sparse-kernel-loss}.
Table~\ref{tab: unsupervised-comparisons} and Figure~\ref{fig: self-supervised-learning-analysis}(e) \& (f) show that using the constraint~\eqref{eq: sparse-kernel-loss} is able to recover better HR videos.
This indicates that better blur kernels are estimated for the construction of the auxiliary supervised constraint.
In addition, we note that the method only using the constraint~\eqref{eq: sparse-kernel-loss} and the self-supervised loss~\eqref{eq: self-supervised-loss-lr} (i.e., ``w/o~\eqref{eq: self-image-loss}" in Table~\ref{tab: unsupervised-comparisons}) does not generate good results, where the PSNR values of this baseline method are at least 2.10dB lower than that of the proposed method.
This demonstrates that using the proposed auxiliary supervised constraint~\eqref{eq: self-image-loss} is able to help video SR when the ground truth videos are not available.

We note that the downsampling operation in the constraint~\eqref{eq: self-supervised-loss-lr} discards most of the pixels in $\mathrm{x}_i$, which means that only a small part of the pixels in $\mathrm{x}_i$ can be constrained.
This may prevent the network $\mathcal{N}_I$ from learning correctly when the ground-truth constraint is not available.
In addition, \eqref{eq: self-supervised-loss-lr} constrains the blur kernel estimation network $\mathcal{N}_k$ and the latent HR frame restoration network $\mathcal{N}_I$ at the same time, which may make the two networks compromise with each other and make the training process unstable.
As the constraint~\eqref{eq: self-image-loss} can effectively regularize the network $\mathcal{N}_I$, we block the constraint~\eqref{eq: self-supervised-loss-lr} from updating the network $\mathcal{N}_I$ with a detach operation, and the constraint~\eqref{eq: self-supervised-loss-lr} is only used to regularize the network $\mathcal{N}_k$.
To demonstrate the effectiveness of this detach operation in~\eqref{eq: self-supervised-loss-lr}, we disable this operation and retrain this baseline method (i.e., ``w/o detach in~\eqref{eq: self-supervised-loss-lr}" in Table~\ref{tab: unsupervised-comparisons}).
It shows that the PSNR values of this baseline method are at least 1.56dB lower than that of the proposed method, which verifies the necessity of this detach operation in the proposed self-supervised learning process.
%

%-------------------------------
\begin{table}[!t]
	\caption{Effectiveness of the temporal information in the constraint~\eqref{eq: self-image-loss} on the REDS4 dataset.}
	\label{tab: lr-flow-com}
	\footnotesize
	\resizebox{0.49\textwidth}{!}{
		\centering
		\begin{tabular}{lcc}
			\toprule
			Methods  &~~~~~~w/o temporal~~~~~~  &~~~~~~w/ temporal (Ours)~~~~~~\\
			\hline
			PNSR  &28.40  &\bf{29.23}\\
			SSIM  &0.8248  &\bf{0.8453}\\
			\bottomrule
		\end{tabular}
	}
\end{table}
%-------------------------------

%-------------------------------
\begin{figure}[!t]\footnotesize
	\centering
	\begin{tabular}{ccc}
		\includegraphics[width=0.32\linewidth]{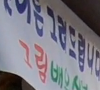} &\hspace{-4mm}
		\includegraphics[width=0.32\linewidth]{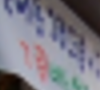} &\hspace{-4mm}
		\includegraphics[width=0.32\linewidth]{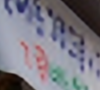}  \\
		(a) HR patch &\hspace{-4mm}  (b) Bicubic & \hspace{-4mm} (c)  ZSSR \\
		\includegraphics[width=0.32\linewidth]{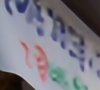} &\hspace{-4mm}
		\includegraphics[width=0.32\linewidth]{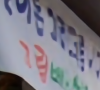} &\hspace{-4mm}
		\includegraphics[width=0.32\linewidth]{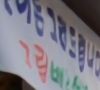}  \\
		(d) MZSR  &\hspace{-4mm}  (e)  w/o temporal &\hspace{-4mm}  (f) Ours \\
	\end{tabular}
	\caption{Effectiveness of the temporal information for video SR.}
	\label{fig: closely-related-methods-analysis}
\end{figure}
%-------------------------------

\subsection{Relations with the unpaired and self-supervised learning image SR methods}

We note that several methods~\cite{zssr,kernelgan} develop effective self-supervised learning methods based on deep neural networks to solve the single image SR problem.
In~\cite{zssr}, Shocher et al. develop an effective image-specific CNN model to solve the single image SR problem which is achieved by the zero-shot learning algorithm.
In the training process, this method first learns the LR-to-HR mapping function from the coarser-resolution one of the input LR image and then applies the learned mapping function to the input LR image for the latent HR image restoration.
This training strategy is similar to our constraint~\eqref{eq: self-image-loss}. However, the proposed constraint~\eqref{eq: self-image-loss} implicitly contains the temporal information which can explore the useful features from adjacent frames for better video SR.
Without using the temporal information, the proposed method would not generate better results.
To verify this, we use the features of original generated auxiliary LR frames $\{\mathrm{L}_i^e\}$ instead of $\{\mathrm{L}_i^{e,w}\}$ in~\eqref{eq: self-image-loss} for comparisons (``w/o temporal" for short).
Table~\ref{tab: lr-flow-com} shows that directly using the features of original generated auxiliary LR frames $\{\mathrm{L}_i^e\}$ instead of $\{\mathrm{L}_i^{e,w}\}$ in~\eqref{eq: self-image-loss} does not generate good videos.
In addition, the ZSSR method does not involve the blur kernel estimation. Although its performance can be significantly improved by using the blur kernels from~\cite{kernelgan}, the super-resolved images would be affected by inaccurate kernels.
In contrast, our method explicitly involves the blur kernel estimation, where blur kernel estimation, the optical flow estimation, and video frame restoration are simultaneously solved in a unified framework.
Thus, it can better reduce the influence of the inaccurate kernel by~\cite{kernelgan}.
The comparisons shown in Figure~\ref{fig: closely-related-methods-analysis}(e) and (f) demonstrate that without using temporal information does not generate clear SR images.

We further note that Maeda~\cite{PseudoSR/cvpr20} develops an effective image SR algorithm based on unpaired data, where the network training does not require the corresponding HR images.
This method needs several networks to generate pseudo supervision from additional exemplar datasets to train the networks.
However, using more networks will accordingly increase the difficulty of the network training.
In contrast, our method does not require any other additional exemplar datasets, and our video degradation constraint does not introduce additional networks, which makes the training process easier than~\cite{PseudoSR/cvpr20}.
Moreover, the method by~\cite{PseudoSR/cvpr20} is designed for single image SR, which cannot be extended to the blind video SR problem directly.
%

%-------------------------------
\begin{table}[!t]
	\caption{Comparisons of estimated blur kernels on the REDS4 dataset in terms of regenerated LR videos.}
	\label{tab: kernels}
	\footnotesize
	\resizebox{0.48\textwidth}{!}{
		\centering
		\begin{tabular}{lcccc}
			\toprule
			Methods  &KernelGAN  &w/o~\eqref{eq: sparse-kernel-loss} \&~\eqref{eq: self-image-loss}  &~~~w/o~\eqref{eq: sparse-kernel-loss}~~~  &~~~Ours~~~\\
			\hline
			PSNR  &33.50  &30.49  &33.95  &\bf{39.62}\\
			SSIM  &0.9721  &0.9421  &0.9752  &\bf{0.9913}\\
			\bottomrule
		\end{tabular}
	}
\end{table}
%-------------------------------

%-------------------------------
%-------------------------------

\subsection{Accuracy of the auxiliary synthesized LR videos}
\label{ssec: analysis-auxiliary-synthesized-LR-videos}

As we do not have any HR videos as the supervision, we propose to estimate blur kernels and optical flow to explore the effective constraint from LR videos based on the image formation model~\eqref{eq: sr-formation-convolution}.
Thus, the estimated blur kernels are mainly used to generate the synthesized LR videos for constructing the auxiliary supervised constraint~\eqref{eq: self-image-loss}, which can regularize the deep model for better HR video restoration.

Table~\ref{tab: kernels} shows that the quality of the synthesized LR videos by using the estimated blur kernels is better than that of the ones generated by the blur kernels~\cite{kernelgan}.
The results also indicate that the proposed method generates better blur kernels.
Although the proposed method does not focus on the accurate recovery of blur kernels, we still visualize the estimated blur kernels in Figure~\ref{fig: kernels}.
It shows that the proposed constraints~\eqref{eq: self-image-loss} and~\eqref{eq: sparse-kernel-loss} is able to improve blur kernel estimation.
%

%-------------------------------
\begin{table}[!t]
	\caption{Evaluations on real scenarios in terms of the average NIQE values.}
	\label{tab: online-finetune}
	\footnotesize
	\resizebox{0.49\textwidth}{!}{
		\centering
		\begin{tabular}{lccc}
			\toprule
			\multirow{2}{*}{Methods}  &\multirow{2}{*}{EDVR~\cite{edvr}}  &\multicolumn{2}{c}{Ours}\\
			\cline{3-4}
			\multicolumn{2}{c}{~} &w/o fine-tuning  &w/ fine-tuning\\
			\hline
			NIQE  &13.688  &13.357  &\bf{12.483}\\
			\bottomrule
		\end{tabular}
	}
\end{table}
%-------------------------------

%-------------------------------
\begin{figure}[!t]\scriptsize
	\centering
	\begin{tabular}{cccc}
		\includegraphics[width=0.24\linewidth,height=0.18\linewidth]{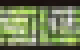} &\hspace{-4mm}
		\includegraphics[width=0.24\linewidth,height=0.18\linewidth]{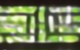} &\hspace{-4mm}
		\includegraphics[width=0.24\linewidth,height=0.18\linewidth]{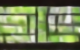} &\hspace{-4mm}
		\includegraphics[width=0.24\linewidth,height=0.18\linewidth]{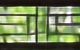}  \\
		\includegraphics[width=0.24\linewidth,height=0.18\linewidth]{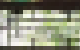} &\hspace{-4mm}
		\includegraphics[width=0.24\linewidth,height=0.18\linewidth]{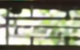} &\hspace{-4mm}
		\includegraphics[width=0.24\linewidth,height=0.18\linewidth]{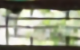} &\hspace{-4mm}
		\includegraphics[width=0.24\linewidth,height=0.18\linewidth]{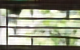}  \\
		(a) LR input  &\hspace{-4mm}  (b) EDVR~\cite{edvr}  &\hspace{-4mm}  (c) w/o fine-tuning &\hspace{-4mm}  (d) w/ fine-tuning \\
	\end{tabular}
	\caption{Effectiveness of the online fine-tuning on real scenarios. }
	\label{fig: online-finetune}
\end{figure}
%-------------------------------

\subsection{Online fine-tuning on real scenarios}

As our method does not require ground truth HR videos as the supervision, real-world degraded videos can be used to fine-tune our method for better generalization. Thus, our method can be fine-tuned online.
To verify this property, we randomly choose 10 real degraded videos from websites to evaluate our method.
As the corresponding ground truth HR videos are not available, we use the non-reference metric NIQE~\cite{NIQE} to evaluate the proposed method.
Table~\ref{tab: online-finetune} shows that our method generates results with better visual perceptual quality than the supervised video SR method~\cite{edvr}.
Moreover, using the proposed self-supervised approach to fine-tune our method with degraded videos is able to generate better results on these real scenarios (see ``w/ fine-tuning" in Table~\ref{tab: online-finetune}).
Figure~\ref{fig: online-finetune} shows some visual results, where our method with fine-tuning using real degraded videos generates better results with finer structures (see Figure~\ref{fig: online-finetune}(d)).
%

%-------------------------------
\begin{table}[!t]
	\caption{Comparisons of the results by the proposed method with supervision of HR videos.}
	\label{tab: supervisio-com}
	\footnotesize
	\resizebox{0.49\textwidth}{!}{
		\centering
		\begin{tabular}{lcc}
			\toprule
			Methods  &~~~~w/ HR supervision~~~~  &~~~~w/o HR supervision (Ours)~~~~\\
			\hline
			PSNR  &27.99  &27.77\\
			SSIM  &0.8346  &0.8184\\
			\bottomrule
		\end{tabular}
	}
\end{table}
%-------------------------------

\subsection{Evaluations of the proposed method using HR videos as supervision}

As the proposed method is designed for the blind video SR problem when the ground truth HR videos are not available, it is interesting to examine whether the proposed method works well if the HR videos are used to supervise the training of the proposed network.
To this end, we use the LR videos and their corresponding HR videos to train the proposed network, where the commonly used $L_1$-norm based loss function is applied to ensure the network output is close to the HR frame.
Table~\ref{tab: supervisio-com} shows that the proposed method without using the HR videos as the supervision generates comparable results compared to the method using the HR videos as the supervision on the SPMCS dataset, which further indicates that the proposed self-supervised learning method is able to solve the blind video SR problem when the ground truth videos are not available.
%

%-------------------------------
\begin{table}[!t]
	\caption{Computational complexity and model size comparisons. The results are obtained on the same machine with the test videos of $720 \times 1280$ pixels.}
	\label{tab: model-size}
	\footnotesize
	\resizebox{0.49\textwidth}{!}{
		\centering
		\begin{tabular}{lccccc}
			\toprule
			Methods  &RCAN  &IKC~\cite{IKC}  & RBPN~\cite{VDBPN/cvpr19}  & EDVR~\cite{edvr}  &Ours\\
			\hline
			Parameters (M)  &15.59  &9.05  &12.77  &20.63  &18.24\\
			Running time (/s)  &0.945  &1.376  &0.709  &0.437  &\bf{0.267}\\
			FLOPs (G)  &919.21  &2535.07  &1245.42  &1480.57  &\bf{754.01}\\
			\bottomrule
		\end{tabular}
	}
\end{table}
%-------------------------------

\subsection{Model size, running time, and computational complexity comparisons}

We further compare the model complexity of our method with state-of-the-art methods in terms of model parameters, running time, and floating point operations (FLOPs).
Table~\ref{tab: model-size} shows that our method is efficient and has a lower FLOPs value.

\subsection{Limitations and future work}

In order to avoid the trivial solutions caused by directly minimizing~\eqref{eq: self-supervised-loss-lr} when there are no ground-truth HR videos as supervision, we first explore the sparse property of the estimated blur kernels and then use the estimated blur kernels and original LR input videos to generate auxiliary paired data for constraining the latent HR video restoration process.
The analysis in Section~\ref{ssec: analysis-auxiliary-synthesized-LR-videos} has demonstrated that the proposed self-supervised learning approach can improve the accuracy of the generated auxiliary paired data.
However, as shown in Figure~\ref{fig: kernels}(b), there is still a certain gap between the estimated blur kernel and the ground-truth blur kernel.
Although the proposed method only focuses on the accuracy of the generated auxiliary paired data but not the estimated blur kernels, future work will study how to estimate more accurate blur kernels so that the performance of the proposed self-supervised method can be further improved.
%

%---------------------------------------------------------
\section{Conclusions}

We have proposed an effective video SR method based on a self-supervised learning method and developed a simple and effective method to generate auxiliary
paired data from the original LR input videos to constrain the network training.
We have introduced an optical flow estimation module to exploit the information from adjacent frames for better HR video restoration.
We have shown that our method also can adopt existing supervised deep video SR models for performance improvement when the ground truth HR videos are not available.
By training the proposed algorithm in an end-to-end fashion, we have shown that it performs favorably against state-of-the-art methods on benchmark datasets and real-world videos.

\bibliographystyle{IEEEtran}
\bibliography{myref}

% Generated by IEEEtran.bst, version: 1.13 (2008/09/30)
\begin{thebibliography}{10}
\providecommand{\url}[1]{#1}
\csname url@samestyle\endcsname
\providecommand{\newblock}{\relax}
\providecommand{\bibinfo}[2]{#2}
\providecommand{\BIBentrySTDinterwordspacing}{\spaceskip=0pt\relax}
\providecommand{\BIBentryALTinterwordstretchfactor}{4}
\providecommand{\BIBentryALTinterwordspacing}{\spaceskip=\fontdimen2\font plus
\BIBentryALTinterwordstretchfactor\fontdimen3\font minus
  \fontdimen4\font\relax}
\providecommand{\BIBforeignlanguage}[2]{{%
\expandafter\ifx\csname l@#1\endcsname\relax
\typeout{** WARNING: IEEEtran.bst: No hyphenation pattern has been}%
\typeout{** loaded for the language `#1'. Using the pattern for}%
\typeout{** the default language instead.}%
\else
\language=\csname l@#1\endcsname
\fi
#2}}
\providecommand{\BIBdecl}{\relax}
\BIBdecl

\bibitem{IKC}
J.~Gu, H.~Lu, W.~Zuo, and C.~Dong, ``Blind super-resolution with iterative
  kernel correction,'' in \emph{Proceedings of IEEE Conference on Computer
  Vision and Pattern Recognition}, 2019, pp. 1604--1613.

\bibitem{zssr}
A.~Shocher, N.~Cohen, and M.~Irani, ````zero-shot" super-resolution using deep
  internal learning,'' in \emph{Proceedings of IEEE Conference on Computer
  Vision and Pattern Recognition}, 2018, pp. 3118--3126.

\bibitem{kernelgan}
S.~Bell-Kligler, A.~Shocher, and M.~Irani, ``Blind super-resolution kernel
  estimation using an internal-gan,'' in \emph{Proceedings of Conference on
  Neural Information Processing Systems}, 2019, pp. 284--293.

\bibitem{VDBPN/cvpr19}
M.~Haris, G.~Shakhnarovich, and N.~Ukita, ``Recurrent back-projection network
  for video super-resolution,'' in \emph{Proceedings of IEEE Conference on
  Computer Vision and Pattern Recognition}, 2019, pp. 3897--3906.

\bibitem{edvr}
X.~Wang, K.~C. Chan, K.~Yu, C.~Dong, and C.~Change~Loy, ``Edvr: Video
  restoration with enhanced deformable convolutional networks,'' in
  \emph{Proceedings of IEEE Conference on Computer Vision and Pattern
  Recognition Workshops}, 2019.

\bibitem{Bayesian/vsr/pami/LiuS14}
C.~Liu and D.~Sun, ``On bayesian adaptive video super resolution,'' \emph{IEEE
  Transactions on Pattern Analysis and Machine Intelligence}, vol.~36, no.~2,
  pp. 346--360, 2014.

\bibitem{huang/videosr/pami18}
Y.~Huang, W.~Wang, and L.~Wang, ``Video super-resolution via bidirectional
  recurrent convolutional networks,'' \emph{IEEE Transactions on Pattern
  Analysis and Machine Intelligence}, vol.~40, no.~4, pp. 1015--1028, 2018.

\bibitem{renjialiao/cvpr15}
R.~Liao, X.~Tao, R.~Li, Z.~Ma, and J.~Jia, ``Video super-resolution via deep
  draft-ensemble learning,'' in \emph{Proceedings of IEEE Conference on
  Computer Vision and Pattern Recognition}, 2015, pp. 531--539.

\bibitem{Kappeler/tmi16}
A.~Kappeler, S.~Yoo, Q.~Dai, and A.~K. Katsaggelos, ``Video super-resolution
  with convolutional neural networks,'' \emph{IEEE Transactions on Medical
  Imaging}, vol.~2, no.~2, pp. 109--122, 2016.

\bibitem{VESPCN}
J.~Caballero, C.~Ledig, A.~P. Aitken, A.~Acosta, J.~Totz, Z.~Wang, and W.~Shi,
  ``Real-time video super-resolution with spatio-temporal networks and motion
  compensation,'' in \emph{Proceedings of IEEE Conference on Computer Vision
  and Pattern Recognition}, 2017, pp. 2848--2857.

\bibitem{ding/liu/iccv17}
D.~Liu, Z.~Wang, Y.~Fan, X.~Liu, Z.~Wang, S.~Chang, and T.~S. Huang, ``Robust
  video super-resolution with learned temporal dynamics,'' in \emph{Proceedings
  of IEEE International Conference on Computer Vision}, 2017, pp. 2526--2534.

\bibitem{taoxin/iccv17}
X.~Tao, H.~Gao, R.~Liao, J.~Wang, and J.~Jia, ``Detail-revealing deep video
  super-resolution,'' in \emph{Proceedings of IEEE International Conference on
  Computer Vision}, 2017, pp. 4482--4490.

\bibitem{DUF}
Y.~Jo, S.~W. Oh, J.~Kang, and S.~J. Kim, ``Deep video super-resolution network
  using dynamic upsampling filters without explicit motion compensation,'' in
  \emph{Proceedings of IEEE International Conference on Computer Vision}, 2018,
  pp. 3224--3232.

\bibitem{TDAN/CVPR2020}
Y.~Tian, Y.~Zhang, Y.~Fu, and C.~Xu, ``Tdan: Temporally-deformable alignment
  network for video super-resolution,'' in \emph{Proceedings of IEEE Conference
  on Computer Vision and Pattern Recognition}, 2020, pp. 3357--3366.

\bibitem{PseudoSR/cvpr20}
S.~Maeda, ``Unpaired image super-resolution using pseudo-supervision,'' in
  \emph{Proceedings of IEEE Conference on Computer Vision and Pattern
  Recognition}, 2020, pp. 291--300.

\bibitem{IBP}
M.~Irani and S.~Peleg, ``Improving resolution by image registration,''
  \emph{Graphical Model and Image Processing}, vol.~53, no.~3, pp. 231--239,
  1991.

\bibitem{vsr/prior/Richard96}
R.~R. Schultz and R.~L. Stevenson, ``Extraction of high-resolution frames from
  video sequences,'' \emph{IEEE Transactions on Image Processing}, vol.~5,
  no.~6, pp. 996--1011, 1996.

\bibitem{shan/sr/sa08}
Q.~Shan, Z.~Li, J.~Jia, and C.~Tang, ``Fast image/video upsampling,'' \emph{ACM
  Transactions on Graphics}, vol.~27, no.~5, pp. 153:1--153:7, 2008.

\bibitem{vsr/prior/Milanfar/tip09}
H.~Takeda, P.~Milanfar, M.~Protter, and M.~Elad, ``Super-resolution without
  explicit subpixel motion estimation,'' \emph{IEEE Transactions on Image
  Processing}, vol.~18, no.~9, pp. 1958--1975, 2009.

\bibitem{SRCNN/ECCV}
C.~Dong, C.~C. Loy, K.~He, and X.~Tang, ``Learning a deep convolutional network
  for image super-resolution,'' in \emph{Proceedings of European Conference on
  Computer Vision}, 2014, pp. 184--199.

\bibitem{VDSR}
J.~Kim, J.~K. Lee, and K.~M. Lee, ``Accurate image super-resolution using very
  deep convolutional networks,'' in \emph{Proceedings of IEEE Conference on
  Computer Vision and Pattern Recognition}, 2016, pp. 1646--1654.

\bibitem{RCAN}
Y.~Zhang, K.~Li, K.~Li, L.~Wang, B.~Zhong, and Y.~Fu, ``Image super-resolution
  using very deep residual channel attention networks,'' in \emph{Proceedings
  of European Conference on Computer Vision}, 2018, pp. 294--310.

\bibitem{TGA/CVPR2020}
T.~Isobe, S.~Li, X.~Jia, S.~Yuan, G.~G. Slabaugh, C.~Xu, Y.-L. Li, S.~Wang, and
  Q.~Tian, ``Video super-resolution with temporal group attention,'' in
  \emph{Proceedings of IEEE Conference on Computer Vision and Pattern
  Recognition}, 2020, pp. 8005--8014.

\bibitem{ziyangma/CVPR15}
Z.~Ma, R.~Liao, X.~Tao, L.~Xu, J.~Jia, and E.~Wu, ``Handling motion blur in
  multi-frame super-resolution,'' in \emph{Proceedings of IEEE Conference on
  Computer Vision and Pattern Recognition}, 2015, pp. 5224--5232.

\bibitem{Tomer/blindsr/iccv13}
T.~Michaeli and M.~Irani, ``Nonparametric blind super-resolution,'' in
  \emph{Proceedings of IEEE International Conference on Computer Vision}, 2013,
  pp. 945--952.

\bibitem{pan2021deep}
J.~Pan, H.~Bai, J.~Dong, J.~Zhang, and J.~Tang, ``Deep blind video
  super-resolution,'' in \emph{Proceedings of IEEE International Conference on
  Computer Vision}, 2021, pp. 4811--4820.

\bibitem{noisetonoise/icml18}
J.~Lehtinen, J.~Munkberg, J.~Hasselgren, S.~Laine, T.~Karras, M.~Aittala, and
  T.~Aila, ``Noise2noise: Learning image restoration without clean data,'' in
  \emph{Proceedings of International Conference on Machine Learning}, 2018, pp.
  2971--2980.

\bibitem{GAN/data/eccv18}
A.~Bulat, J.~Yang, and G.~Tzimiropoulos, ``To learn image super-resolution, use
  a gan to learn how to do image degradation first,'' in \emph{Proceedings of
  European Conference on Computer Vision}, 2018, pp. 187--202.

\bibitem{PWCNet}
D.~Sun, X.~Yang, M.-Y. Liu, and J.~Kautz, ``Pwc-net: Cnns for optical flow
  using pyramid, warping, and cost volume,'' in \emph{Proceedings of IEEE
  Conference on Computer Vision and Pattern Recognition}, 2018, pp. 8934--8943.

\bibitem{CDVD/deblur/cvpr20}
J.~Pan, H.~Bai, and J.~Tang, ``Cascaded deep video deblurring using temporal
  sharpness prior,'' in \emph{Proceedings of IEEE Conference on Computer Vision
  and Pattern Recognition}, 2020, pp. 3040--3048.

\bibitem{edsr}
B.~Lim, S.~Son, H.~Kim, S.~Nah, and K.~M. Lee, ``Enhanced deep residual
  networks for single image super-resolution,'' in \emph{Proceedings of IEEE
  Conference on Computer Vision and Pattern Recognition}, 2017, pp. 1132--1140.

\bibitem{shi2016real}
W.~Shi, J.~Caballero, F.~Husz{\'a}r, J.~Totz, A.~P. Aitken, R.~Bishop,
  D.~Rueckert, and Z.~Wang, ``Real-time single image and video super-resolution
  using an efficient sub-pixel convolutional neural network,'' in
  \emph{Proceedings of IEEE Conference on Computer Vision and Pattern
  Recognition}, 2016, pp. 1874--1883.

\bibitem{adam}
D.~P. Kingma and J.~Ba, ``Adam: {A} method for stochastic optimization,''
  \emph{CoRR}, vol. abs/1412.6980, 2014.

\bibitem{reds}
S.~Nah, S.~Baik, S.~Hong, G.~Moon, S.~Son, R.~Timofte, and K.~M. Lee, ``Ntire
  2019 challenge on video deblurring and super-resolution: Dataset and study,''
  in \emph{Proceedings of IEEE Conference on Computer Vision and Pattern
  Recognition Workshops}, 2019, pp. 1996--2005.

\bibitem{MZSR}
J.~W. Soh, S.~Cho, and N.~I. Cho, ``Meta-transfer learning for zero-shot
  super-resolution,'' in \emph{Proceedings of IEEE Conference on Computer
  Vision and Pattern Recognition}, 2020, pp. 3513--3522.

\bibitem{NIQE}
A.~Mittal, R.~Soundararajan, and A.~C. Bovik, ``Making a “completely blind”
  image quality analyzer,'' \emph{IEEE Signal Processing Letters}, vol.~20,
  no.~3, pp. 209--212, 2012.

\end{thebibliography}

%\begin{IEEEbiography}[{\includegraphics[width=1in,height=1.25in,clip,keepaspectratio]{./figures/authors/hrbai}}]{Haoran Bai} is currently pursuing the Ph.D. degree with the School of Computer Science and Engineering, Nanjing University of Science and Technology, China. His research interests include image/video super-resolution, deblurring, dehazing, and other restoration tasks.
%\end{IEEEbiography}
%
%
%\begin{IEEEbiography}[{\includegraphics[width=1in,height=1.25in,clip,keepaspectratio]{./figures/authors/jspan}}]{Jinshan Pan} received the Ph.D. degree in computational mathematics from the Dalian University of Technology, China, in 2017. He is a Professor at the School of Computer Science and Engineering, Nanjing University of Science and Technology, China. His research interests include deblurring, image/video analysis and enhancement, and related vision problems. He is a member of the IEEE.
%\end{IEEEbiography}
%
%% You can push biographies down or up by placing
%% a \vfill before or after them. The appropriate
%% use of \vfill depends on what kind of text is
%% on the last page and whether or not the columns
%% are being equalized.
%
%%\vfill
%
%% Can be used to pull up biographies so that the bottom of the last one
%% is flush with the other column.
%\enlargethispage{-5in}

% that's all folks
\end{document}